\pdfoutput=1

\documentclass[11pt]{article}

\usepackage[]{ACL2023}
\usepackage{graphicx}
\usepackage{caption}
\usepackage{subcaption}
\usepackage{times}
\usepackage{latexsym}
\usepackage{multirow}
\usepackage{algorithm}
\usepackage{algpseudocode}
\usepackage{multicol}
\usepackage{amsmath}
\usepackage{amssymb}
\usepackage{algpascal}
\usepackage{inconsolata}
\usepackage{letltxmacro}
\usepackage{booktabs}
\usepackage{caption}
\usepackage[T1]{fontenc}

\usepackage[utf8]{inputenc}

\usepackage{microtype}

\usepackage{colortbl,xcolor,xspace}
\usepackage{soul}
\usepackage{subcaption}

\usepackage{caption}
\usepackage{comment}
\usepackage{subcaption}

\newcommand{\mconer}{\textsc{MultiCoNER}\xspace}
\newcommand{\langid}[1]{\texttt{#1}}
\title{ACLM: A Selective-Denoising based Generative Data Augmentation Approach for Low-Resource Complex NER}

\usepackage{fdsymbol}
\newcommand\blfootnote[1]{%
  \begingroup
  \renewcommand\thefootnote{}\footnote{#1}%
  \addtocounter{footnote}{-1}%
  \endgroup
}

\author{Sreyan Ghosh$^{\spadesuit*}$ \quad Utkarsh Tyagi$^{\spadesuit*}$ \quad Manan Suri$^{\clubsuit}$\quad Sonal Kumar$^{\spadesuit}$ \\
\bf S Ramaneswaran$^{\varheartsuit}$ \quad \bf Dinesh Manocha$^{\spadesuit}$  \\
        $^{\spadesuit}$University of Maryland, College Park, USA, \\
        $^{\clubsuit}$NSUT Delhi, India,
         $^{\varheartsuit}$NVIDIA, Bangalore, India \\
         \texttt{\{sreyang, utkarsht, sonalkum, dmanocha\}@umd.edu} \\
         \texttt{manansuri27@gmail.com, ramanr@nvidia.com}}
  
\begin{document}
\maketitle
\begin{abstract}

Complex Named Entity Recognition (NER) is the task of detecting linguistically complex named entities in low-context text. In this paper, we present ACLM (\underline{\textbf{A}}ttention-map aware keyword selection for \underline{\textbf{C}}onditional \underline{\textbf{L}}anguage \underline{\textbf{M}}odel fine-tuning), a novel data augmentation approach, based on conditional generation, to address the data scarcity problem in low-resource complex NER. ACLM alleviates the context-entity mismatch issue, a problem existing NER data augmentation techniques suffer from and often generates incoherent augmentations by placing complex named entities in the wrong context. ACLM builds on BART and is optimized on a novel text reconstruction or denoising task - we use \emph{selective masking} (aided by attention maps) to retain the named entities and certain \emph{keywords} in the input sentence that provide contextually relevant additional knowledge or hints about the named entities. Compared with other data augmentation strategies, ACLM can generate more diverse and coherent augmentations preserving the true word sense of complex entities in the sentence. We demonstrate the effectiveness of ACLM both qualitatively and quantitatively on monolingual, cross-lingual, and multilingual complex NER across various low-resource settings. ACLM outperforms all our neural baselines by a significant margin (1\%-36\%). In addition, we demonstrate the application of ACLM to other domains that suffer from data scarcity (e.g., biomedical). In practice, ACLM generates more effective and factual augmentations for these domains than prior methods.\footnote{Code: https://github.com/Sreyan88/ACLM} \blfootnote{$^*$These authors contributed equally to this work.}

\end{abstract}

\section{Introduction}
\label{sec:intro}

Named Entity Recognition (NER) is a fundamental task in Natural Language Processing (NLP) that aims to detect various types of named entities (NEs) from text. Recently, there has been considerable progress in NER using neural learning methods that achieve state-of-the-art (SOTA) performance~\cite{wang2020automated,zhou2021learning} on well-known benchmark datasets, including CoNLL 2003 ~\cite{conll} and OntoNotes~\cite{schwartz2012improving}. However, these datasets are designed to evaluate the performance on detecting ``relatively easy'' NEs like~\emph{proper names} (e.g., people such as “Barack Obama,” locations such as “New York,” or organizations such as “IBM”) in well-formed, context-rich text that comes from news articles~\cite{augenstein2017generalisation}. On the other hand, complex NER benchmarks like  MultiCoNER~\cite{malmasi2022multiconer} present several contemporary challenges in NER, including short low-context texts with emerging and semantically ambiguous complex entities (e.g., movie names in online comments) that reduce the performance of SOTA methods previously evaluated only on the existing NER benchmark datasets. Our experiments reveal that the performance of the current SOTA NER method~\cite{zhou2021learning} (previously evaluated only on the CoNLL 2003 dataset) drops by 23\% when evaluated on MultiCoNER and 31.8\% when evaluated on a low-resource setting with just 500 training samples (more details in Table \ref{tab:performance_simple_ner}). Thus, we emphasize that research on building systems that can effectively detect complex NEs in the text is currently understudied in the field of NLP.

In the past, researchers have made several attempts at building  supervised approaches to detect complex and compositional noun phrase entities in sentences~\cite{doddington2004automatic,biggio2010entity,magnolini2019use}. However, the scarcity of annotated training data for building effective systems has always been a challenge. Data augmentation has been shown to be an effective solution for low-resource NER~\cite{ding2020daga,liu2021mulda,zhou2022melm}. In practice, though these systems perform well and generate coherent augmentations on common NER benchmark datasets with easy proper noun NEs, they fail to be effective for complex NER, often generating incoherent augmentations. We first argue that certain types of complex NEs follow specific linguistic patterns and appear only in specific contexts (examples in Appendix \ref{fig:linguistic}), and augmentations that do not follow these patterns impede a NER model from learning such patterns effectively. This sometimes also leads to augmentations with context-entity mismatch, further hurting the learning process. For e.g., unlike proper names, substituting complex NEs from other sentences in the corpus or replacing them with synonyms \cite{dai2020analysis} often leads to augmentations where the NE does not fit into the new context (e.g., swapping proper names across sentences might still keep the sentence coherent \emph{but} swapping the name of a book with a movie (both \emph{creative work} entity) \emph{or} the name of a football team with a political party (both \emph{group} entity) makes it incoherent). Fine-tuning pre-trained language models (PLMs), similar to prior-work \cite{ding2020daga,liu2021mulda,zhou2022melm}, fail to generate new context around complex NEs or completely new NEs with the desired linguistic patterns due to low-context sentences and the lack of existing knowledge of such linguistically complex NEs (examples in Fig. \ref{fig:nlg-examples}). This leads to in-coherent augmentations and poses a severe problem in knowledge-intensive tasks like bio-medical NER, where non-factual augmentations severely hurt learning.
Our experiments also reveal that introducing new context patterns around NEs proves to be a more effective data augmentation technique for complex NER than diversifying NEs (ACLM vs. MELM in Table \ref{tab:performance_mono_cross}).
\vspace{1mm}





{\noindent \textbf{Main Results:}} To overcome the aforesaid problems, we formulate data augmentation as a conditional generation task and propose ACLM, a conditional text generation model that generates augmentation samples by introducing new and diverse context patterns around a NE. ACLM builds on BART \cite{lewis-etal-2020-bart} and is fine-tuned on a modification of the text reconstruction from corrupted text task, a common denoising-based PLM pre-training objective. In contrast to other PLM pre-training strategies, which randomly mask a portion of the text for corruption, our modified objective is based on \emph{selective masking}, wherein we mask all other words in the sentence except the NEs and a small percentage of \emph{keywords} related to the NEs. We refer to this corrupted sentence as a \emph{template}, and it serves as input to the model for both the training and generation phases. These keywords are other non-NE tokens in the sentence that provide contextually relevant additional knowledge or hints to BART about the complex NEs without the need of retrieving knowledge from any external sources. We select these keywords using attention maps obtained from a transformer model fine-tuned on the NER task, and they help the PLM overcome the problem where it might not possess enough knowledge about a semantically ambiguous complex NE (example in Fig. \ref{fig:nlg-examples}). Training ACLM on this modified objective allows us to generate diverse, coherent, factual, and high-quality augmentations given templates. We also propose \emph{mixner}, a novel algorithm that mixes two templates during the augmentation generation phase and boosts the diversity of augmentations. Our primary contributions are as follows:

\begin{itemize}
    \item We propose ACLM, a novel data augmentation framework specially designed for low-resource complex NER. Compared with previous methods in the literature, ACLM effectively alleviates the context-entity mismatch problem by preserving the true sense of semantically ambiguous NEs in augmentations. Additionally, to accompany ACLM, we propose \emph{mixner}, which boosts the diversity of ACLM generations. 
    \item We qualitatively and quantitively show the benefits of ACLM for monolingual, cross-lingual, and multilingual complex NER across various low-resource settings on the MultiCoNER dataset. Our proposed ACLM outperforms all other baselines in literature by a significant margin (1\%-36\%) and generates more diverse, coherent, and high-quality augmentations compared to them.
    \item We perform extensive experiments to study the application of ACLM in three other domains, including science and medicine. ACLM outperforms all our baselines in these domains (absolute gains in the range of 1\%-11\%) and generates more factual augmentations.
\end{itemize}

\section{Background and Related Work}
\label{sec:back_rel}

{\noindent \textbf{Complex NER Background:}} Complex NER is a relatively understudied task in the field of NLP. Building on insights from \citet{augenstein2017generalisation}, we discuss key reasons behind high performance on common NER benchmark datasets and try to understand why modern SOTA NER algorithms do not work well on complex NER benchmarks: (1) \textbf{Context}: Most of the common benchmark datasets are curated from articles in the news domain. This gives them several advantages, including rich context and surface features like proper punctuation and capitalized nouns, all of which are major drivers of success in these datasets \cite{mayhew2019ner}. In contrast, for entity recognition beyond news text, like search queries or voice commands, the context is less informative and lacks surface features \cite{guo2009named,carmel2014erd}; (2) \textbf{Entity Complexity}: Data from news articles contain \emph{proper names} or “easy” entities with simple syntactic structures, thus allowing pre-trained models to perform well due to their existing knowledge of such entities. On the other hand, complex NEs like movie names are syntactically ambiguous and linguistically complex and which makes Complex NER a difficult task~\cite{ashwini2014targetable}. Examples of such entities include noun phrases (e.g., Eternal Sunshine of the Spotless Mind), gerunds (e.g., Saving Private Ryan), infinitives (e.g., To Kill a Mockingbird), or full clauses (e.g., Mr. Smith Goes to Washington); (3) \textbf{Entity Overlap}: Models trained on these common benchmark datasets suffer from memorization effects due to the large overlap of entities between the train and test sets. Unseen and emerging entities pose a huge challenge to complex NER \cite{bernier2020hardeval}.

{\noindent \textbf{Complex NER:}} Prior work has mostly focused on solving the entity complexity problem by learning to detect complex nominal entities in sentences~\cite{magnolini2019use,meng2021gemnet,fetahu-etal-2022-dynamic,chen2022ustc}. Researchers have often explored integrating external knowledge in the form of gazetteers for this task. Gazetteers have also proven to be effective for low-resource NER~\cite{rijhwani2020soft}. GemNet~\cite{meng2021gemnet}, the current SOTA system for complex NER, conditionally combines the contextual and gazetteer features using a Mixture-of-Experts (MoE) gating mechanism. However, gazetteers are difficult to build and maintain and prove to be ineffective for complex NER due to their limited entity coverage and the nature of unseen and emerging entities in complex NER.
\vspace{1mm}

{\noindent \textbf{Data Augmentation for Low-Resource NER:}} Data Augmentation to handle data scarcity for low-resource NLP is a well-studied problem in the literature and is built on word-level modifications, including simple synonym replacement strategies \cite{wei-zou-2019-eda}, or more sophisticated learning techniques like LSTM-based language models \cite{kobayashi-2018-contextual}, Masked Language Modeling (MLM) using PLMs \cite{kumar2020data}, auto-regressive PLMs \cite{kumar2020data}, or constituent-based tagging schemes \cite{zhou-etal-2019-dual}. However, most of these methods, though effective for classification tasks, suffer from token-label misalignment when applied to token-level tasks such as NER and might require complex pre-processing steps \cite{bari2020uxla,zhong2021time}. One of the first works to explore effective data augmentation for NER replaces NEs with existing NEs of the same type or replaces tokens in the sentence with one of their synonyms retrieved from WordNet \cite{dai-adel-2020-analysis}. Following this, many neural learning systems were proposed that either modify the Masked Language Modelling (MLM) training objective using PLMs \cite{zhou2022melm,liu22low} or use generative language modeling with LSTM LMs \cite{ding2020daga} or mBART \cite{liu2021mulda}, to produce entirely new sentences from scratch. However, all these systems were designed for low-resource NER on common benchmark datasets and failed to generate effective augmentations for low-resource complex NER with semantically ambiguous and complex entities. 
\vspace{1mm}

\begin{figure*}[t]
\centering
\includegraphics[width=2\columnwidth]{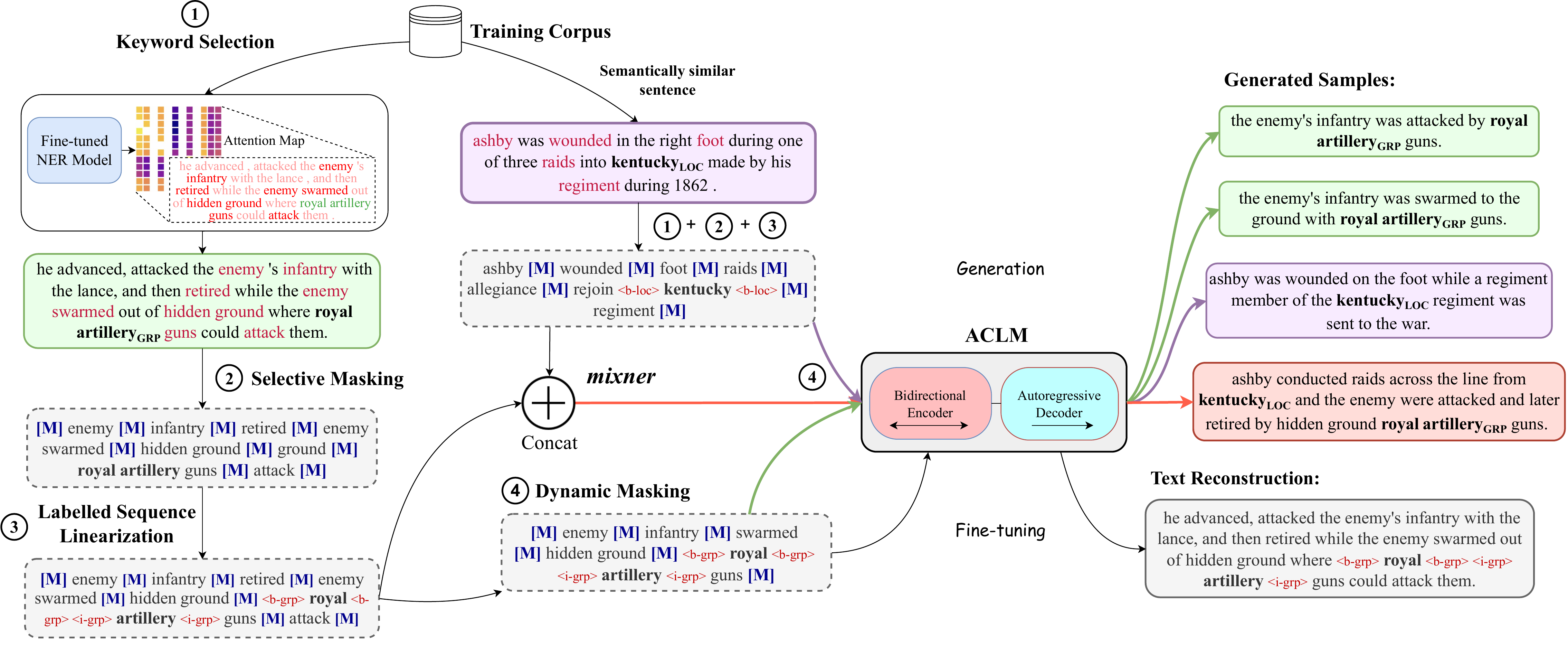}
\caption{\small \textbf{Overview of ACLM:} ACLM follows a 4-step template creation process, which serves as an input to the model during fine-tuning and generation. \textcircled{\raisebox{-0.9pt}{1}} \textbf{Keyword Selection:} The most important keywords (in red) associated with the NEs (in bold) in the sentence is first extracted using attention maps obtained from a fine-tuned NER model. \textcircled{\raisebox{-0.9pt}{2}} \textbf{Selective Masking:} All words except the NEs and the keywords obtained from the previous step is replaced with mask tokens $\lbrack$\textbf{M}$\rbrack$. \textcircled{\raisebox{-0.9pt}{3}} \textbf{Labeled Sequence Linearization:} Label tokens are added before and after each entity in the sentence. \textcircled{\raisebox{-0.9pt}{4}} \textbf{Dynamic Masking:} The template goes through further masking where a small portion of the keywords are dynamically masked at each training iteration. While generation we also apply \emph{mixner}, which randomly joins two templates after \textcircled{\raisebox{-0.9pt}{3}} and before \textcircled{\raisebox{-0.9pt}{4}}. Post generating augmentations with ACLM, the generated augmentations are concatenated with the gold data and used to fine-tune our final NER model.}
\label{fig:aclm}
\end{figure*}

\section{Methodology}
\label{sec:methodology}

In this section, we give an overview of our approach.
Fig. \ref{fig:aclm} represents the entire workflow of our ACLM data augmentation framework. A sentence is first passed through a fine-tuned XLM-RoBERTa fine-tuned on only gold data to generate the attention map for each token in the sentence. This attention map is then used to selectively mask the sentence and create a template. This template is then used as an input to optimize the model on the text reconstruction objective for fine-tuning ACLM: the model is asked to reconstruct the entire original sentence from only the content in the template. While generating augmentations, ACLM follows the same template generation process in addition to adding two templates through \emph{mixner}, which we discuss in detail in Section \ref{subsec:generation}.


\subsection{Template Creation}
\label{sec:template_gen}
To corrupt a sentence and create a template, we follow a 4-step process described below:
\vspace{1mm}

{\noindent 1. \textbf{Keyword Selection}}: For each sentence in our training corpus, we first obtain a set of non-NE tokens in the sentence that are most attended to by its NEs. We call these tokens \emph{keywords}. For our research, we consider a non-NE token as a keyword if the NEs in the sentence contextually depend on them the most. We measure contextual dependency between NE and non-NE tokens using attention scores from attention maps extracted from a transformer-based NER model fine-tuned only on gold data. We hypothesize that attention heads in a transformer when fine-tuned for NER, formulated as a token-level tagging task, tend to pay the highest attention to the most contextually relevant tokens around it. Thus, formally put, consider a sentence with a total of $T$ tokens comprised of $t_{other}$ non-NE and $t_{entity}$ NE tokens. Our primary aim is to find the top $p\%$ of $t_{other}$ tokens, which we call keywords. To calculate the total attention score that each token in the sentence assigns to each other token, we sum up the attention scores across each of the heads in the transformer network and across the last $a$ layers ($a$ = 4 in our case). Different heads in different layers tend to capture different properties of language, and taking the average attention scores across the last 4 layers ensures that diverse linguistic relations are taken into account while choosing the keywords (e.g., syntactic, semantic, etc.). This also makes the keyword selection process more robust, as in low-resource conditions the attention maps may be noisy, and the NEs might not be focusing on the right context always. Additionally, the choice of just the last four layers is inspired by the fact that the lower layers have very broad attention and spend at most 10\% of their attention mass on a single token \cite{clark2019does}. Note $t_{entity}$ might be comprised of (1) multiple contiguous tokens forming an individual NE and (2) multiple such individual NEs. To handle the first case, inspired from \citet{clark2019does}, we sum up the attention scores over all the individual tokens in the NE. For the second case, we find $t_{attn}$ for each individual NE and take a set union of tokens in these $t_{attn}$. Thus, as an extra pre-processing step, to improve robustness, we also ignore punctuations, stop words, and other NEs from the top $p\%$ of $t_{other}$ tokens to obtain our final keywords. We provide examples of templates in Appendix \ref{sec:templates}.
\vspace{1mm}

{\noindent 2. \textbf{Selective Masking}}: After selecting the top $p\%$ of $t_{other}$ tokens in the sentence as keywords, we now have $K$ non-NE keyword tokens and $E$ entity tokens. To create the template, we now substitute each non-NE token not belonging to the $K$ with the mask token and remove contiguous mask tokens.
\vspace{1mm}

{\noindent 3. \textbf{Labeled Sequence Linearization}}: After we have our initial template, inspired by \citet{zhou2022melm}, we perform labeled sequence linearization to explicitly take label information into consideration during fine-tuning and augmentation generation. Similar to \citet{zhou2022melm}, as shown in Figure \ref{fig:aclm}, we add label tokens before and after each entity token and treat them as the normal context in the sentence. Additionally, these label tokens before and after each NE provide boundary supervision for NEs with multiple tokens.
\vspace{1mm}

{\noindent 4. \textbf{Dynamic Masking}}: Post labeled sequence linearization, our template goes through further masking wherein we dynamically mask a small portion of the $K$ keywords during each iteration of training and generation. To be precise, we first sample a dynamic masking rate $\varepsilon$ from a Gaussian distribution $\mathcal{N}(\mu,\,\sigma^{2})$, where the Gaussian variance $\sigma$ is set to 1/$K$. Next, we randomly sample tokens from the $K$ keywords in the sentence according to the masking rate $\varepsilon$ and replace this with mask tokens, followed by removing consecutive mask tokens. At every round of generation, dynamic masking helps boost 1) context diversity by conditioning ACLM generation on different templates with a different set of keywords and 2) length diversity by asking ACLM to infill a different number of mask tokens.

\subsection{Fine-tuning ACLM}
As discussed earlier, ACLM is fine-tuned on a novel text reconstruction from corrupted text task wherein the created templates serve as our corrupted text and ACLM learns to recover the original text from the template. Text reconstruction from the corrupted text is a common denoising objective that PLMs like BART and BERT are pre-trained on. For this work, we use it as our fine-tuning objective and differ from other existing pre-training objectives by our \emph{selective masking} strategy for creating templates. 


\subsection{Data Generation}
\label{subsec:generation}
Post fine-tuning on the text reconstruction task, we utilize ACLM to generate synthetic data for data augmentation. For each sentence in the training dataset, we apply steps 1-4 in the Template Creation pipeline for $R$ rounds to randomly corrupt the sentence and obtain a template which is then passed through the fine-tuned ACLM model to generate a total of $R\times$ augmented training samples. Additionally, to boost diversity, during auto-regressive generation, we randomly sample the next word from the \emph{top-k} most probable words and choose the most probable sequence with beam search.
\vspace{1mm}

\begin{figure}[t]
\centering
\includegraphics[width=0.9\columnwidth]{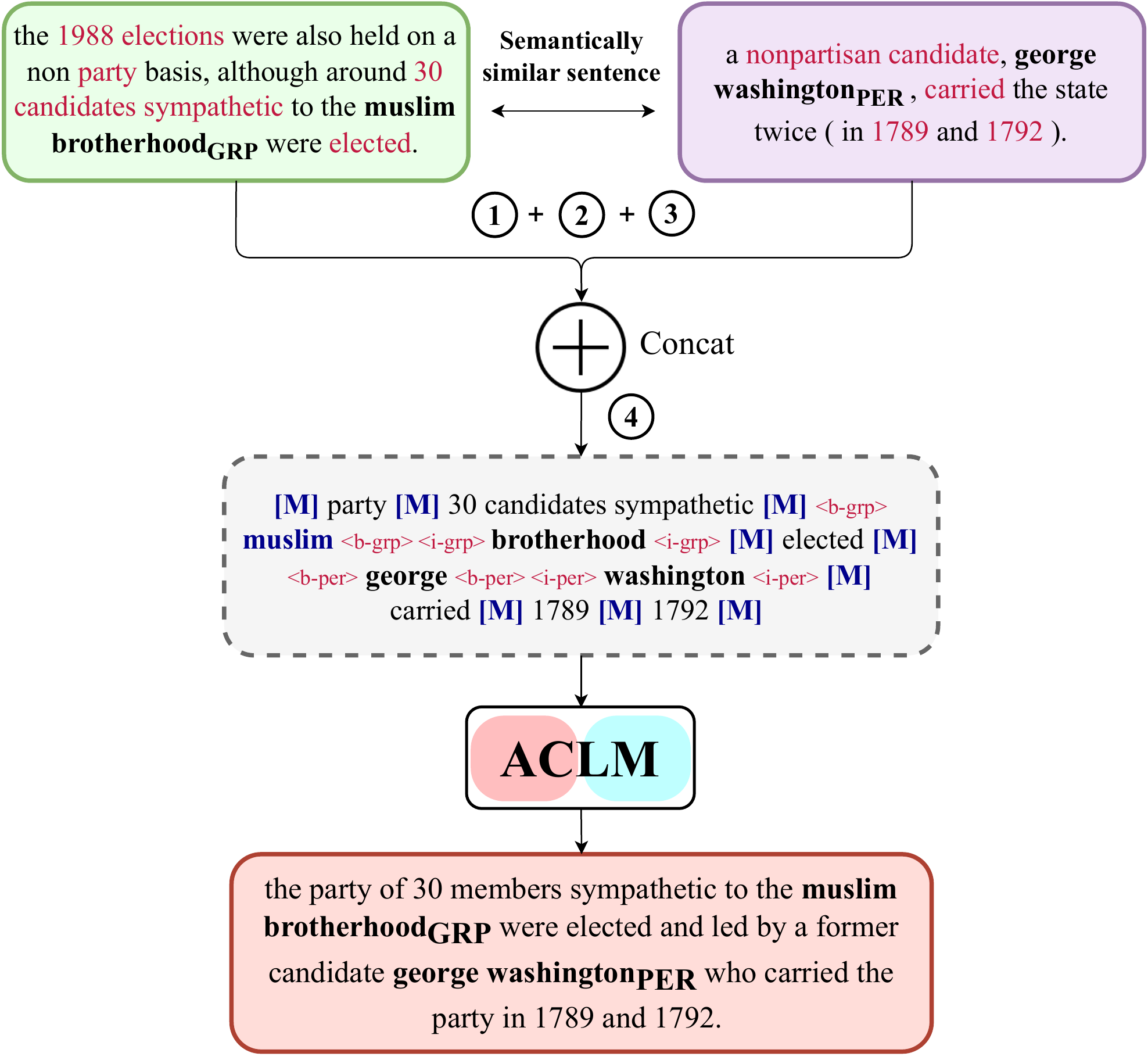}
\caption{\small \textbf{Overview of \emph{mixner}:} During the augmentation generation process, for a particular sentence in the training dataset, we retrieve another semantically similar sentence and concatenate them before step  \textcircled{\raisebox{-0.9pt}{4}} of the template creation process. This merged template is then passed through ACLM to generate diverse augmentations that incorporate semantics and NEs from both sentences.}
\label{fig:mixner}
\end{figure}

{\noindent \textbf{\emph{mixner}:}} During the $R$ rounds of augmentation on our training dataset, we propose the use of \emph{mixner}, a novel template mixing algorithm that helps ACLM generate diverse sentences with new context and multiple NEs in the sentence. More specifically, given the template for any arbitrary sentence $a$ in the training set in step 3 of the template creation process, we retrieve the template for another sentence $b$ that is semantically similar to $a$ and join both the templates before passing on the template to step 4. We show examples of sentences generated with \emph{mixner} in Fig. \ref{fig:nlg-examples} and Section \ref{subsec:aug_examples}. Note that we apply \emph{mixner} only in the generation step and not during fine-tuning.

As mentioned earlier, to retrieve $b$ from the training set, we randomly sample a sentence from the \emph{top-k} sentences with the highest semantic similarity to $a$. To calculate semantic similarity between each sentence in the training set, we first take the embedding $e$ for each sentence from a multi-lingual Sentence-BERT \cite{reimers-2019-sentence-bert} and then calculate semantic similarity by:

\begin{equation}
    {\operatorname{sim}}(e_i,e_j) = \frac{e_i \cdot e_j}{\left\|e_i\right\|\left\|e_j\right\|}
\end{equation}


where $\operatorname{sim(.)}$ is the cosine similarity between two embeddings , and $i,j \in N$ where $i \neq j$, and $N$ is the size if the training set. Additionally, we don't apply \emph{mixner} on all rounds $R$ but sample a probability $\gamma$ from a Gaussian distribution $\mathcal{N}(\mu,\,\sigma^{2})$ and only apply \emph{mixner} if $\gamma$ crosses a set threshold $\beta$.

\subsubsection{Post-Processing}
\label{subsubsec:postprocessing}
As a post-processing step, we remove augmentations similar to the original sentence and also the extra label tokens added in the labeled sequence linearization step. Finally, we concatenate the augmented data with the original data to fine-tune our NER model.
\setlength{\tabcolsep}{4pt}
{\renewcommand{\arraystretch}{1.2}%
\begin{table*}[t]
\centering
\small
\resizebox{2.05\columnwidth}{!}{
\begin{tabular}{clccccccccccc||ccccc}
\hline
\hline
    &   & \multicolumn{10}{c}{\textsc{Monolingual}} & & \multicolumn{5}{c}{\textsc{Cross-Lingual}} \\
\textbf{\#Gold} & \textbf{Method} & \textbf{En} & \textbf{Bn} & \textbf{Hi} & \textbf{De} & \textbf{Es} & \textbf{Ko }& \textbf{Nl} & \textbf{Ru} & \textbf{Tr} & \textbf{Zh} & \textbf{Avg} & $\textbf{En}\rightarrow{\text{\textbf{Hi}}}$ & $\textbf{En}\rightarrow{\text{\textbf{Bn}}}$ & $\textbf{En}\rightarrow{\text{\textbf{De}}}$ & $\textbf{En}\rightarrow{\text{\textbf{Zh}}}$ & \textbf{Avg}  \\
\hline
     & Gold-only & 29.36 & 14.49 & 18.80 & 37.04 & 36.30 & 12.76 & 38.78 & 23.89 & 24.13 & 14.18 & 24.97 & 16.36 & 12.15 & 29.71 & 0.31 & 14.63 \\
     & LwTR & 48.60 & 20.25 & 29.95 & 48.38 & 44.08 & 35.09 & 43.00 & 39.22 & 30.58 & 27.70 & 36.68 & 32.36 & \cellcolor{magenta!20}\textbf{24.59} & 46.05 & 2.11 & 26.28 \\
 & DAGA & 16.24 & 5.87 & 10.40 & 32.44 & 27.78 & 19.28 & 15.44 & 11.14 & 16.17 & 10.33 & 16.51 & 4.54 & 3.28 & 14.21 & 0.13 & 5.54 \\
100  & MELM & 40.12 & 6.22 & 27.84 & 43.94 & 37.45 & 34.10 & 37.82 & 32.38 & 20.13 & 25.11 & 30.51 & 26.37 & 20.33 & 34.32 & 2.71 & 20.93 \\
& ACLM \emph{only entity} & 14.06 & 17.55 & 19.60 & 29.72 & 38.10 & 31.57 & 38.47 & 27.40 & 35.62 & 26.34 & 27.84 & 21.72 & 16.55 & 30.93 & 1.58 & 17.69 \\
& ACLM \emph{random} & 43.59 & 20.13 & 28.04 & 45.83 & 42.27 & 33.64 & 41.82 & 38.20 & 36.79 & 25.99 & 35.63 & 29.68 & 21.64 & 45.27 & 3.05 & 24.91\\
     & \textbf{ACLM \emph{(ours)}} &  \cellcolor{magenta!20}\textbf{48.76}  &  \cellcolor{magenta!20}\textbf{23.09}  &  \cellcolor{magenta!20}\textbf{33.53}  &  \cellcolor{magenta!20}\textbf{48.80}  &  \cellcolor{magenta!20}\textbf{44.14}  &  \cellcolor{magenta!20}\textbf{38.35}  &  \cellcolor{magenta!20}\textbf{46.22}  &  \cellcolor{magenta!20}\textbf{39.48}  &  \cellcolor{magenta!20}\textbf{37.20} &  \cellcolor{magenta!20}\textbf{35.12}  & \cellcolor{magenta!20}\textbf{39.47}  & \cellcolor{magenta!20}\textbf{32.52} & 23.91 & \cellcolor{magenta!20}\textbf{46.48} & \cellcolor{magenta!20}\textbf{3.58} & \cellcolor{magenta!20}\textbf{26.62}\\
     \hline
     & Gold-only & 51.83 & 19.31 & 33.68 & 49.62 & 45.16 & 42.51 & 47.83 & 31.55 & 26.76 & 32.34 & 38.06 & 36.90 & 27.44 & 48.70 & 3.76 & 29.20\\
     & LwTR & 52.88 & 23.85 & 34.27 & 50.31 & 47.01 & 42.77 & 52.01 & 40.18 & 35.92 & 30.57 & 40.98 & 40.07 & 32.36 & 48.95 & 6.04 & 31.85\\
 & DAGA & 33.30 & 17.12 & 19.58 & 35.10 & 33.56 & 26.50 & 38.04 & 29.83 & 23.35 & 25.66 & 28.20 & 18.92 & 14.37 & 29.32 & 1.79 & 16.10 \\
200 & MELM & 47.83 &5.47 & 29.67 & 45.85 & 42.08 & 36.62 & 49.47 & 41.84 & 31.25 & 32.27 & 36.24 & 27.55 & 18.80 & 41.10 & 6.21 & 23.41  \\
& ACLM \emph{only entity} & 50.06 & 25.58 & 37.78 & 50.95 & 48.21 & 43.39 & 48.46 & 34.87 & 34.92 &28.20& 40.24 & 30.76 & 22.53 & 44.17 & 6.50 & 25.99 \\
& ACLM \emph{random}& 52.69 & 35.26 & 39.83 & 51.14 & 48.70 & 42.19 & 48.71 & 39.68 & 37.26 & 34.22 & 42.96 & 36.52 & 27.19 & 47.73 & 7.12 & 29.64 \\
     & \textbf{ACLM \emph{(ours)}} & \cellcolor{magenta!20}\textbf{54.99} & \cellcolor{magenta!20}\textbf{38.39} & \cellcolor{magenta!20}\textbf{40.55} & \cellcolor{magenta!20}\textbf{53.36} & \cellcolor{magenta!20}\textbf{49.57} & \cellcolor{magenta!20}\textbf{44.32} & \cellcolor{magenta!20}\textbf{53.19} & \cellcolor{magenta!20}\textbf{43.97} & \cellcolor{magenta!20}\textbf{39.71} & \cellcolor{magenta!20}\textbf{39.31} & \cellcolor{magenta!20}\textbf{45.74} & \cellcolor{magenta!20}\textbf{45.22} & \cellcolor{magenta!20}\textbf{36.64} & \cellcolor{magenta!20}\textbf{54.51} & \cellcolor{magenta!20}\textbf{8.55}  &
    \cellcolor{magenta!20}\textbf{36.23} \\
     \hline
     & Gold-only & 55.51 & 34.6 & 38.66 & 55.95 & 51.52 & 48.57 & 50.97 & 45.14 & 38.83 & 38.84 & 45.86 & 35.93 & 25.64 & 50.13 & 7.23 & 29.73\\
     & LwTR & 56.97 & 35.42 & 37.83 & 55.91 & 54.74 & 49.36 & 56.10 & 46.82 & 39.00 & 38.55 & 47.07 & 43.14 & 34.60 & 51.61 & 11.40 & 35.19 \\
 & DAGA & 44.62 & 22.36 & 24.30 & 43.02 & 42.77 & 36.23 & 47.11 & 30.94 & 30.84 & 33.79 & 35.60 & 26.50 & 21.52 & 37.89 & 4.82 & 22.68 \\
500 & MELM &52.57 & 9.46 & 31.57 & 53.57 & 46.40 & 45.01 & 51.90 & 46.73 & 38.26 & 39.64 & 41.51 & 34.97 & 27.17 & 44.31 & 7.31 & 28.44\\
& ACLM \emph{only entity} & 57.55 & 35.69 & 35.82 & 56.15 & 53.64 & 50.20 & 53.07 & 46.40 & 41.58 & 38.65 & 46.87 & 35.48 & 29.37 & 49.10 & 7.99 & 30.48\\
& ACLM \emph{random} & 57.92 & 38.24 & 39.33 & 57.14 & 53.24 & 49.81 & 55.06 & 48.27 & 42.22 & 40.55 & 48.18 & 41.72 & 32.16 & 52.27 & 13.63 & 34.95\\
& \textbf{ACLM \emph{(ours)}} & \cellcolor{magenta!20}\textbf{58.31} & \cellcolor{magenta!20}\textbf{40.26} & \cellcolor{magenta!20}\textbf{41.48} & \cellcolor{magenta!20}\textbf{59.35} & \cellcolor{magenta!20}\textbf{55.69} & \cellcolor{magenta!20}\textbf{51.56} & \cellcolor{magenta!20}\textbf{56.31} & \cellcolor{magenta!20}\textbf{49.40} & \cellcolor{magenta!20}\textbf{43.57} & \cellcolor{magenta!20}\textbf{41.23} & \cellcolor{magenta!20}\textbf{49.72} & \cellcolor{magenta!20}\textbf{44.36} & \cellcolor{magenta!20}\textbf{35.59} & \cellcolor{magenta!20}\textbf{54.04} & \cellcolor{magenta!20}\textbf{16.27} &
\cellcolor{magenta!20}\textbf{37.57} \\
     \hline
     & Gold-only & 57.22 & 30.20 & 39.55 & 60.18 & 55.86 & 53.39 & 60.91 & 49.93 & 43.67 & 43.05 & 44.40 & 43.44 & 33.27 & 54.61 & 5.34 & 34.17\\
     & LwTR  &  59.10  & 39.65   & 43.90   & 61.28 & 57.29  & 51.37 & 59.25   & 52.04   &  44.33  &  43.71  &  51.19  & 43.32   & 33.74  & 53.32  & 7.38 & 34.44 \\
 & DAGA & 50.24 & 32.09 & 35.02 & 51.45 & 49.47 & 42.41 & 51.88 & 41.56 & 33.18 & 39.51 & 42.68 & 33.12 & 26.22 & 42.13 & 5.15 & 26.65 \\
1000 & MELM & 53.48 & 6.88 & 37.02 & 58.69 & 52.43 & 50.50 & 56.25 & 48.99 & 36.83 & 38.88 & 44.00 & 35.23 & 25.64 & 46.50 & 8.22 & 28.90 \\
& ACLM \emph{only entity} & 55.46 & 38.13 & 41.84 & 60.05 & 56.99 & 53.32 & 58.22 & 50.17 & 45.11 & 39.62 & 49.89 & 37.38 & 29.77 & 41.10 & 6.49 & 28.69\\
& ACLM \emph{random} & 58.87 & 41.00 & 46.27 & 61.19 & 57.29 & 53.61 & 59.52 & 52.77 & 45.01 & 43.60 & 51.91 & 43.96 & 34.14 & 53.37 & 7.25 & 34.68\\   
     & \textbf{ACLM \emph{(ours)}} & \cellcolor{magenta!20}\textbf{60.14} & \cellcolor{magenta!20}\textbf{42.42} & \cellcolor{magenta!20}\textbf{48.20} & \cellcolor{magenta!20}\textbf{63.80} & \cellcolor{magenta!20}\textbf{58.33} & \cellcolor{magenta!20}\textbf{55.55} & \cellcolor{magenta!20}\textbf{61.22} & \cellcolor{magenta!20}\textbf{54.31} & \cellcolor{magenta!20}\textbf{48.23} & \cellcolor{magenta!20}\textbf{45.19} & \cellcolor{magenta!20}\textbf{53.74} & \cellcolor{magenta!20}\textbf{44.59} & \cellcolor{magenta!20}\textbf{35.70} & \cellcolor{magenta!20}\textbf{56.74} & \cellcolor{magenta!20}\textbf{8.94} &
     \cellcolor{magenta!20}\textbf{36.49} \\
     \hline
     \hline
\end{tabular}
}

    \caption{\small Results of monolingual (Left) and cross-lingual (Right) low-resource complex NER. For cross-lingual experiments, we take English as the source language. ACLM obtains absolute average gains in the range of 1\% - 22\% over our baselines.}
    \label{tab:performance_mono_cross}
\end{table*}}

\section{Experiments and Results}

\subsection{Dataset}
\label{sec:dataset}

All our experiments were conducted on the MultiCoNER dataset \cite{malmasi2022multiconer}, a large multilingual dataset for complex NER. MultiCoNER covers 3 domains, including Wiki sentences, questions, and search queries, across 11 distinct languages. The dataset represents contemporary challenges in NER discussed in Section \ref{sec:back_rel} and is labeled with six distinct types of entities: \textbf{person}, \textbf{location}, \textbf{corporation}, \textbf{groups} (political party names such as \emph{indian national congress}), \textbf{product} (consumer products such as \emph{apple iPhone 6}), and \textbf{creative work} (movie/song/book titles such as \emph{on the beach}). We conduct experiments on a set of 10 languages $\mathbb{L}$ where $\mathbb{L}$ = \{English (\textbf{En}), Bengali (\textbf{Bn}, Hindi (\textbf{Hi}), German (\textbf{De}), Spanish (\textbf{Es}), Korean (\textbf{Ko}), Dutch (\textbf{Nl}), Russian (\textbf{Ru}), Turkish (\textbf{Tr}), Chinese (\textbf{Zh})\}. Language-wise dataset statistics can be found in Table \ref{tab:data_stats_multiconner}. We would also like to highlight that the number of sentences in MultiCoNER test sets ranges from \textbf{133,119 - 217,887}, which is much higher than test sets of other existing NER datasets. For more details on the dataset, we refer our readers to \citet{malmasi2022multiconer}. For monolingual and cross-lingual low-resource experiments, we  perform iterative stratified sampling over all the sentences by using the entity classes in a sample as its target label across four low-resource settings (100, 200, 500, and 1000). We downsample the development set accordingly. For multi-lingual experiments, we combine all the data sampled for our monolingual settings. We evaluate all our systems and baselines on the original MultiCoNER test sets. We report micro-averaged F1 scores averaged across 3 runs for 3 different random seeds.

\algdef{SE}[FOR]{NoDoFor}{EndFor}[1]{\algorithmicfor\ #1}{\algorithmicend\ \algorithmicfor}
\algdef{SE}[IF]{NoThenIf}{EndIf}[1]{\algorithmicif\ #1}{\algorithmicend\ \algorithmicif}
\renewcommand{\algorithmiccomment}[1]{\hfill$\triangleright${#1}}
\begin{algorithm}[t]
\scriptsize
\caption{ACLM: Our proposed augmentation framework}
\label{alg:algo1}
    \begin{algorithmic}
    \State $\textbf{\text{Given }} \text{training set } \mathbb{D}_{\text{train}}, \text{ and PLM } \mathcal{L}$
    \State $\mathbb{D}_{masked} \gets \emptyset, \mathbb{D}_{aug} \gets \emptyset$
    \NoDoFor \textbf{for} {${\{X,Y\}} \in \mathbb{D}_{train}$} \textbf{do}
    \Comment{Training Loop}
    \State $t_{other}, t_{entity} \gets X$
    \State $K \gets Top \text{ } p\% \text{ } of \textsc{AttnMap}(t_{other}$)
    \Comment{Keyword Selection}
      \State $\tilde{X} \leftarrow \textsc{GenTemplate}(X, \{t_{other}\}-\{K\})$
      \Comment{Selective Masking}
      \State $\tilde{X} \leftarrow \textsc{Linearize}(\tilde{X}, Y)$
      \State $\mathbb{D}_{\text{masked }} \leftarrow \mathbb{D}_{\text{masked }} \cup\{\tilde{X}\}$
    \EndFor \textbf{end for}
    \NoDoFor \textbf{for} {${\{X,Y\}} \in \mathbb{D}_{masked}$} \textbf{do}
    \State $\tilde{X} \leftarrow \textsc{DynamicMask}(X, \eta)$
      \Comment{Dynamic Masking}
    \State $\mathcal{L}_{finetune } \leftarrow \textsc{Finetune}(\mathcal{L}, \tilde{X})$
    \Comment{Fine-tune ACLM}
    \EndFor $\textbf{\text{end for }}$
    \NoDoFor $\textbf{\text{for }} \{{{X,Y\}} \in \mathbb{D}_{train}} \textbf{\text{ do}}$
    \Comment{Generation Loop}
    \State $\textbf{\text{repeat }} \mathcal{R} \textbf{\text{ times}}$:
            \State $\hspace{1em}\tilde{X} \leftarrow \textsc{GenTemplate}(X, \{t_{other}\}-\{K\})$
            \Comment{Selective masking}
            \State $\hspace{1em}\tilde{X} \leftarrow \textsc{Linearize}(\tilde{X}, Y)$
            \Comment{Labeled Sequence Linearization}
            \State $\hspace{1em}\tilde{X} \leftarrow \textsc{DynamicMask}(\tilde{X}, \mu)$
            \Comment{Dynamic Masking}
            \State $\hspace{1em}X_{aug} \leftarrow \textsc{GenAug}(\mathcal{L}_{finetune}(\tilde{X})), \text{ } if \text{ } \gamma < \beta$
            \State $\hspace{1em}X_{augmix} \leftarrow \textsc{Mixner}(\mathcal{L}_{finetune}(\tilde{X})), \text{ } if \text{ } \gamma > \beta$
            \State $\hspace{1em}\mathbb{D}_{aug } \leftarrow \mathbb{D}_{aug} \cup \{X_{aug}\}\cup \{X_{augmix}\}$
    \EndFor $\textbf{\text{end for }}$
    \State $\mathbb{D}_{aug} \leftarrow \textsc{PostProcess}(\mathbb{D}_{aug})$
    \Comment{Post-processing}
    \State \textbf{return} $\mathbb{D}_{train} \cup \mathbb{D}_{aug}$
    \end{algorithmic}
\end{algorithm}

\subsection{Experimental Setup}

{\noindent \textbf{ACLM.}} We use mBart-50-large \cite{tang2020multilingual} with a condition generation head to fine-tune ACLM. We fine-tune ACLM for 10 epochs using Adam optimizer \cite{kingma2014adam} with a learning rate of $1e^{-5}$ and a batch size of 32.
\vspace{1mm}

{\noindent \textbf{NER.}} We use XLM-RoBERTa-large with a linear head as our NER model. Though the field of NER has grown enormously, in this paper, we adhere to the simplest formulation and treat the task as  a token-level classification task with a BIO tagging scheme. We use the Adam optimizer to optimize our model, set the learning rate to $1e^{-2}$, and train with a batch size of 16. The NER model is trained for 100 epochs, and the model with the best performance on the dev set is used for testing. 
\vspace{1mm}


{\noindent \textbf{Hyper-parameter Tuning.}} For template creation during fine-tuning and generation, we set the selection rate $p$ and the Gaussian $\mu$ to be 0.3 and 0.5, respectively. The number of augmentation rounds $R$ is set as 5. For \emph{mixner} we set Gaussian $\mu$ and $\beta$ to be 0.5 and 0.7, respectively. All hyper-parameters are tuned on the development set with grid search. More details can be found in Appendix \ref{sec:hyper}.

\subsection{Baselines}
To prove the effectiveness of our proposed ACLM, we compare it with several strong NER augmentation baselines in the literature. In this sub-section, we briefly describe each of these baselines. All baselines were run for $R$ rounds.



\vspace{0.5mm}

{\noindent \textbf{Gold-Only.}} The NER model is trained using only gold data from the MultiCoNER dataset without any augmentation.
\vspace{0.5mm}

{\noindent \textbf{Label-wise token replacement (LwTR).}\cite{dai-adel-2020-analysis}} A token in a sentence is replaced with another token with the same label; the token is randomly selected from the training set.
\vspace{0.5mm}

{\noindent \textbf{DAGA.}}\cite{ding2020daga} Data Augmentation with a Generation Approach (DAGA) proposes to train a one-layer LSTM-based recurrent neural network language model (RNNLM) by maximizing the probability for the next token prediction with linearized sentences. During generation, they use random sampling to generate entirely new sentences with only the $\lbrack\textbf{BOS}\rbrack$ token fed to the model.

\vspace{0.5mm}

{\noindent \textbf{MulDA.}}\cite{liu2021mulda} Multilingual Data Augmentation Framework (MulDA) builds on DAGA and trains a pre-trained mBART model on next token prediction with linearized sentences for generation-based multilingual data augmentation. For a fair comparison, we replace mBART in MulDA with mBART-50.
\vspace{0.5mm}

{\noindent \textbf{MELM.}}\cite{zhou2022melm} Masked Entity Language Modeling (MELM) proposes fine-tuning a transformer-encoder-based PLM on linearized labeled sequences using masked language modeling. MELM outperforms all other baselines and prior-art on low-resource settings on the CoNLL 2003 NER dataset across four languages in mono-lingual, cross-lingual, and multi-lingual settings.
\vspace{0.5mm}

{\noindent \textbf{ACLM \emph{random}.}} We train and infer ACLM with templates created with randomly sampled \emph{keywords} instead of taking \emph{keywords} with high attention scores. This baseline proves the effectiveness of our \emph{keyword} selection algorithm which provides NEs in the template with rich context.
\vspace{0.5mm}

{\noindent \textbf{ACLM \emph{only entity}.}} We train and infer ACLM with templates created with only linearized entities and no \emph{keywords}. This baseline proves the effectiveness of additional context in our templates.
\vspace{0.5mm}


\subsection{Experimental Results}

{\noindent{\textbf{Monolingual Complex NER.}}} Table \ref{tab:performance_mono_cross} compares the performance of all our baselines with ACLM on the MultiCoNER test sets under various low-resource settings for 10 languages. As clearly evident, ACLM outperforms all our baselines in all settings by consistently achieving the best results in all individual languages. Moreover, ACLM improves over our neural baselines (MELM and DAGA) by a significant margin (absolute gains in the range of 1.5\% - 22\% across individual languages). Although LwTR performs better than ACLM in rare instances, we emphasize that (1) LwTR generates nonsensical, incoherent augmentations, (discussed further in Section \ref{subsec:aug_examples}) and (2) Based on a learning-based paradigm, ACLM shows bigger margins to LwTR at slightly higher gold training samples (200 and 500) which we acknowledge is a reasonable size in real-world conditions.
\vspace{1mm}

{\renewcommand{\arraystretch}{1.1}%
\begin{table}[t]
\centering
\resizebox{\columnwidth}{!}{
\begin{tabular}{lllllllllllll}
\hline\hline
\textbf{\#Gold}   & \textbf{Method} & \textbf{En} & \textbf{Bn} & \textbf{Hi} & \textbf{De} & \textbf{Es} & \textbf{Ko }& \textbf{Nl} & \textbf{Ru} & \textbf{Tr} & \textbf{Zh}& \textbf{Avg}  \\ 
\hline
\multirow{5}{*}{100 $\times$10}    &Gold-Only  & 56.21   & 35.66   &  42.16  &  55.71  &  54.98  &  45.14  &  57.48  &  46.13  &  44.40  & 30.72 & 46.86\\
                        &LwTR   &  55.65  &  38.47  &  43.44  & 54.71   & 53.95   & 44.78   &  56.50  & 46.93   & 45.41   & 31.56 & 47.14\\
                        &MulDA  &  46.87  &  29.25  &  34.52  &  45.92  &  45.55  &  33.91  &  48.21  &  38.65  & 35.56   & 27.33 & 38.58\\
                        &MELM   & 53.27   & 23.43   & 41.55   & 48.17   & 51.28  & 39.23   & 51.37   & 45.73   & 41.97   & 30.67 & 42.67\\
                        &\textbf{ACLM \emph{(ours)}} &  \cellcolor{magenta!20}\textbf{58.74}  & \cellcolor{magenta!20}\textbf{41.00}   & \cellcolor{magenta!20}\textbf{46.22}   &  \cellcolor{magenta!20}\textbf{59.13}  &  \cellcolor{magenta!20}\textbf{56.93}  & \cellcolor{magenta!20}\textbf{51.22}   &  \cellcolor{magenta!20}\textbf{60.30}  &  \cellcolor{magenta!20}\textbf{50.26}  &  \cellcolor{magenta!20}\textbf{49.32}  & \cellcolor{magenta!20}\textbf{40.93} & \cellcolor{magenta!20}\textbf{51.40}\\

\hline
\multirow{5}{*}{200 $\times$10}    &Gold-Only  &  58.67  &  39.84  &  46.34  &  59.65  & 58.50   &  50.70  &  60.79  &  51.66  & 47.12   & 40.98 & 51.42\\
                        &LwTR   &  51.78  & 35.93   & 38.87   & 52.73   & 51.59   &  42.55  & 54.49   &  43.99  & 41.23   & 35.19 & 44.83\\
                        &MulDA  &  48.89  &   31.45 & 36.76   &  48.41  &  48.30  &  39.78  &  51.09  &   42.01 &  35.98  & 31.65 & 41.43\\
                        &MELM  & 52.53   & 24.27   & 40.10   & 49.69   & 52.42   & 43.56   & 47.28   & 44.35   & 40.62   & 34.28 & 47.45 \\
                        &\textbf{ACLM \emph{(ours)}} &  \cellcolor{magenta!20}\textbf{59.75}  &  \cellcolor{magenta!20}\textbf{42.61}  &  \cellcolor{magenta!20}\textbf{48.52}  & \cellcolor{magenta!20}\textbf{61.49}   &  \cellcolor{magenta!20}\textbf{59.05}  &  \cellcolor{magenta!20}\textbf{53.46}  & \cellcolor{magenta!20}\textbf{61.59}   & \cellcolor{magenta!20}\textbf{53.34}   &  \cellcolor{magenta!20}\textbf{49.96}  & \cellcolor{magenta!20}\textbf{44.72} & \cellcolor{magenta!20}\textbf{53.45}\\

\hline
\multirow{5}{*}{500 $\times$10}    &Gold-Only  &  61.10  & 40.94   & 48.20   &  61.67  &  59.84  &  54.56  &  62.36  &  53.33  & 48.77   & 45.82 & 53.66\\
                        &LwTR   & 59.09   & 38.37   &  43.80  & 59.37   & 57.76   & 50.38   & 60.42   &  51.00  &  46.53  & 42.87 & 50.96\\
                        &MulDA  &  51.79  &  30.67  &   35.79 &   51.87 &      50.92& 43.08  &  53.95  & 44.61   & 38.86   & 36.72 & 43.83\\
                        &MELM  & 58.67    & 26.17   & 41.88    & 53.05   & 57.26   & 51.97   & 61.49   & 43.73   & 40.22   & 40.12 & 47.66\\
                        &\textbf{ACLM \emph{(ours)}} &  \cellcolor{magenta!20}\textbf{62.32}  & \cellcolor{magenta!20}\textbf{43.79}   &  \cellcolor{magenta!20}\textbf{50.32}  &  \cellcolor{magenta!20}\textbf{63.94}  & \cellcolor{magenta!20}\textbf{62.05}   &  \cellcolor{magenta!20}\textbf{56.82}  & \cellcolor{magenta!20}\textbf{64.41}   &  \cellcolor{magenta!20}\textbf{55.09}  & \cellcolor{magenta!20}\textbf{51.83}   & \cellcolor{magenta!20}\textbf{48.44} & \cellcolor{magenta!20}\textbf{55.90}\\
\hline
\multirow{5}{*}{1000 $\times$10}    &Gold-Only  & 64.14   &  43.28  & 50.11   &  66.18  & 63.17   & 57.31   & 65.75   &  56.94  &  51.17  & 49.77 & 57.78\\
                        &LwTR   & 61.67   & 39.90   & 45.28   & 63.13   &  60.21  &  53.43  &  63.37  &  54.07  & 48.38   & 45.36 & 53.48\\
                        &MulDA  &  56.35  &   33.73 &  40.71  &  56.90  &  55.35  &  48.42  &  58.39  & 49.25   &  42.06  & 40.19 & 48.14\\
                        &MELM  & 61.55   & 30.27   & 42.61   & 61.05   & 61.87   & 55.71   & 63.17   & 53.00   & 48.48   & 44.71 & 52.24\\
                        &\textbf{ACLM \emph{(ours)}} & \cellcolor{magenta!20}\textbf{64.50}   &  \cellcolor{magenta!20}\textbf{46.59}  &  \cellcolor{magenta!20}\textbf{52.14}  &  \cellcolor{magenta!20}\textbf{67.65}  & \cellcolor{magenta!20}\textbf{64.02}   &  \cellcolor{magenta!20}\textbf{59.09}  & \cellcolor{magenta!20}\textbf{67.03}   &  \cellcolor{magenta!20}\textbf{57.82}  &  \cellcolor{magenta!20}\textbf{53.25}  & \cellcolor{magenta!20}\textbf{50.60} & \cellcolor{magenta!20}\textbf{58.27}\\

\hline\hline
\end{tabular}
}
\caption{\small Results of multi-lingual low-resource complex NER. ACLM obtains absolute gains in the range of 1\% - 21\%.}
\label{tab:multi_all}
\end{table}}

{\noindent{\textbf{Cross-lingual Complex NER.}}} We also study the cross-lingual transferability of a NER model trained on a combination of gold and generated augmentations. Thus, we evaluated a model, trained on \textbf{En}, on 4 other languages, including \textbf{Hi}, \textbf{Bn}, \textbf{De}, and \textbf{Zh} in a zero-shot setting. ACLM outperforms our neural baselines by a significant margin (absolute gains in the range of 1\% - 21\%). None of these systems perform well in cross-lingual transfer to \textbf{Zh} which was also observed by \cite{hu2021investigating}.
\vspace{1mm}

{\noindent{\textbf{Multi-lingual Complex NER.}}} Table \ref{tab:multi_all} compares the performance of all our baselines with ACLM on the MultiCoNER test sets under various multilingual low-resource settings. As clearly evident, ACLM outperforms all our baselines by a significant margin (absolute gains in the range of 1\%-21\% across individual languages). All our baselines, including our Gold-Only baseline, also perform better than their monolingual counterparts which demonstrates the effectiveness of multi-lingual fine-tuning for low-resource complex NER.


\section{Further Analysis}
\subsection{Generation Quality}
{\noindent \textbf{Quantitative Analysis.}} Table \ref{tab:quant_aug} compares augmentations from various systems on the quantitative measures of perplexity and diversity. Perplexity \cite{jelinek1977perplexity} is a common measure of text fluency, and we measure it using GPT2 \cite{radford2019language}. We calculate 3 types of diversity metrics: for Diversity-E and Diversity-N, we calculate the average percentage of new NE and non-NE words in the generated samples compared with the original samples, respectively. For Diversity-L, we calculate the average absolute difference between the number of tokens in generated samples and the original samples. ACLM achieves the lowest perplexity and highest non-NE and length diversity compared with other baselines. NE diversity in ACLM is achieved with \emph{mixner} where ACLM fairs well compared to MELM which just replaces NEs. LwTR achieves the highest perplexity, thereby reaffirming that it generates incoherent augmentations.



{\setlength{\tabcolsep}{2pt}
{\renewcommand{\arraystretch}{1.1}%
\begin{table}[t]
    \tiny
    \centering
    \resizebox{1.0\columnwidth}{!}{
    \begin{tabular}{clcccc}
    \hline\hline
        \textbf{\#Gold} & \textbf{Method} & \textbf{Perplexity($\downarrow$)} & \textbf{Diversity-E($\uparrow$)} & \textbf{Diversity-N($\uparrow$)} & \textbf{Diversity-L($\uparrow$)}\\
        \hline
         \multirow{3}{*}{200}& LwTR & 137.01 & 30.72 & 16.46 & 0.0 \\
        & MELM & 83.21 & \cellcolor{magenta!20}\textbf{94.85} & 0.0 & 0.0 \\
        & \textbf{ACLM \emph{(ours)}} & \cellcolor{magenta!20}\textbf{80.77} & 35.64 & \cellcolor{magenta!20}\textbf{22.48} & \cellcolor{magenta!20}\textbf{5.67} \\ 
         \hline
         \multirow{3}{*}{500}& LwTR & 129.349 & 30.07 & 16.22 & 0.0 \\
        & MELM & 82.31 & \cellcolor{magenta!20}\textbf{94.37} & 0.0 & 0.0 \\
        & \textbf{ACLM \emph{(ours)}} & \cellcolor{magenta!20}\textbf{57.68} & 44.12 & \cellcolor{magenta!20}\textbf{41.16} & \cellcolor{magenta!20}\textbf{5.82} \\
         \hline
         \multirow{3}{*}{1000}& LwTR & 131.20 & 29.85 & 16.55 & 0.0\\
        & MELM & 82.64 & \cellcolor{magenta!20}\textbf{95.13} & 0.0 & 0.0 \\
        & \textbf{ACLM \emph{(ours)}} & \cellcolor{magenta!20}\textbf{62.00} & 50.10 & \cellcolor{magenta!20}\textbf{34.84} & \cellcolor{magenta!20}\textbf{5.40}\\
         \hline
    \end{tabular}}
    \caption{\small Quantitative evaluation of generation quality from various systems on the measures of perplexity and diversity. Diversity-E, N, and L stand for Entity, Non-Entity, and Length, respectively.}
    \label{tab:quant_aug}
\end{table}}}

\begin{figure*}[t]
    \includegraphics[width=0.98\textwidth]{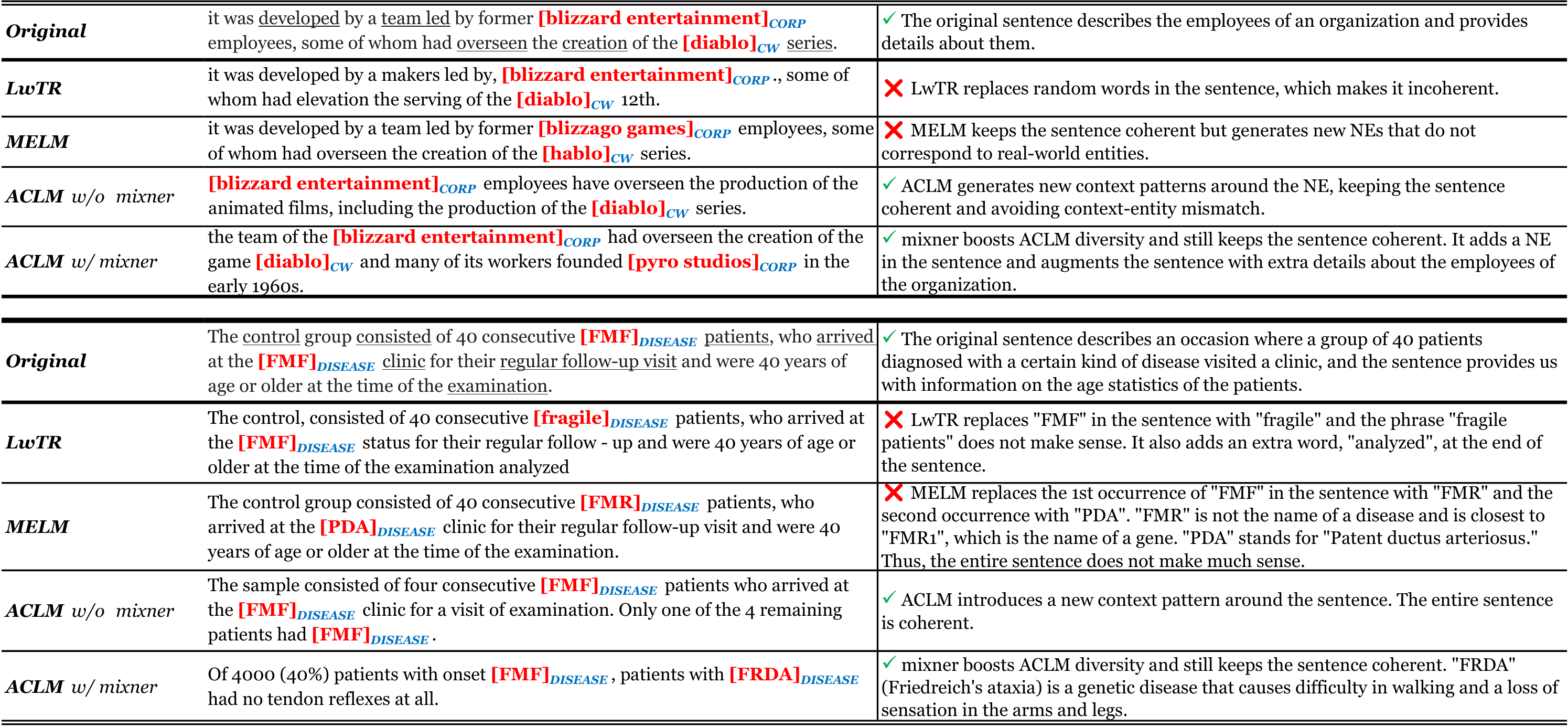}
    \caption{\small Examples of augmentations generated with different methods (Left) and explanation (Right). Words in \textcolor{red}{red} are Named Entities, and words \underline{underlined} in the \emph{Original} sentence are identified ACLM \emph{keywords}. ACLM generates much more diverse, detailed, and coherent augmentations, which maintain factuality and also prove to be more effective. Generation diversity is further amplified with \emph{mixner}.}
  \label{fig:nlg-examples}
\end{figure*}

{\noindent \textbf{Qualtitative Analysis.}} Fig. \ref{fig:nlg-examples} illustrates the superiority of augmentations generated by ACLM when compared with our other baselines. As clearly evident, while MELM generates just minor changes in NEs, augmentations produced by LwTR often tend to be nonsensical and incoherent. On the other hand, ACLM generates meaningful and diverse sentences around NEs, which is further boosted with \emph{mixner}. We provide examples in Appendix \ref{subsec:aug_examples}.

\subsection{Application to other domains}
\label{subsec:transferability}

To evaluate the transferability of ACLM to other domains, we evaluate ACLM on 4 more datasets beyond MultiCoNER. These datasets include CoNLL 2003 \cite{conll} (news), BC2GM \cite{smith2008overview} (bio-medical), NCBI Disease \cite{dougan2014ncbi} (bio-medical) and TDMSci \cite{houyufang2021eacl} (science). Table \ref{tab:other_datsets} compares our baselines with ACLM across 2 low-resource settings on all 4 datasets. ACLM outperforms all our baselines on all settings except LwTR on CoNLL 2003. This occurs because LwTR generates a large variety of effective augmentations with NE replacement on easy entities in CoNLL 2003. The results demonstrate the effectiveness of ACLM over diverse domains, including domains with an acute scarcity of data (bio-medical). Additionally, we also emphasize that ACLM produces more factual augmentations and, unlike our other baselines, avoids context-entity mismatch, which makes the NER model store wrong knowledge in data-sensitive domains. We show samples of generated augmentations in Fig. \ref{tab:quant_aug} and Appendix \ref{subsec:aug_examples}.




 {\renewcommand{\arraystretch}{1.1}%
 \begin{table}[t]
    \tiny
    \centering
    \begin{tabular}{clccccc}
    \hline\hline
        \textbf{\#Gold} & \textbf{Method} & \textbf{CoNLL} & \textbf{BC2GM} & \textbf{NCBI} & \textbf{TDMSci} & \textbf{Avg} \\
        \hline
         \multirow{5}{*}{200} & Gold-Only & 79.11 & 50.01 & 72.92 & 47.20 & 62.31\\
         & LwTR & \cellcolor{magenta!20}\textbf{82.33} & 52.78 & 72.15 & 51.65 & 64.73\\
         & DAGA & 76.23 & 47.67 & 71.14 & 48.03 & 60.77\\
         & MELM & 77.10 &54.05  & 70.12 & 46.07 & 61.83\\
         & \textbf{ACLM \emph{(ours)}} & 82.14 & \cellcolor{magenta!20}\textbf{58.48} & \cellcolor{magenta!20}\textbf{74.27} & \cellcolor{magenta!20}\textbf{56.83} & \cellcolor{magenta!20}\textbf{67.93}\\
         \hline
         \multirow{5}{*}{500} & Gold-Only & 84.82 & 55.56 & 75.75 & 47.04 &65.79\\
         & LwTR & \cellcolor{magenta!20}\textbf{85.08} & 60.46 & 78.97 & 60.74 &71.31\\
         & DAGA & 81.82 & 51.23 & 78.09 & 57.66 & 67.20\\
         & MELM & 83.51 & 56.83 & 75.11 & 57.80 & 68.31\\
         & \textbf{ACLM \emph{(ours)}} & 84.26 & \cellcolor{magenta!20}\textbf{62.37} & \cellcolor{magenta!20}\textbf{80.57} & \cellcolor{magenta!20}\textbf{61.77} & \cellcolor{magenta!20}\textbf{72.24}\\
         \hline
    \end{tabular}
    \caption{\small Comparison of NER results on datasets from various different domains including news, science, and bio-medical.}
    \label{tab:other_datsets}
\end{table}}

\section{Conclusion}
In this paper, we propose ACLM, a novel data augmentation framework for low-resource complex NER. ACLM is fine-tuned on a novel text reconstruction task and is able to generate diverse augmentations while preserving the NEs in the sentence and their original word sense. ACLM effectively alleviates the context-entity mismatch problem and generates diverse, coherent, and high-quality augmentations that prove to be extremely effective for low-resource complex NER. Additionally, we also show that ACLM can be used as an effective data augmentation technique for low-resource NER in the domains of medicine and science due to its ability to generate extremely reliable augmentations.




\section*{Limitations}
We list down some potential limitations of ACLM: 1) PLMs are restricted by their knowledge to generate entirely new complex entities due to their syntactically ambiguous nature. Adding to this, substituting complex NEs in existing sentences leads to context-entity mismatch. Thus, as part of future work, we would like to explore if integrating external knowledge into ACLM can help generate sentences with new complex entities in diverse contexts. 2) We do not conduct experiments in the language Farsi from the MultiCoNER dataset as neither mBart-50-large nor XLM-RoBERTa-large was pre-trained on this language. 3) The use of mBart-50-large for generation also restricts ACLM from being transferred to code-switched settings, and we would like to explore this as part of future work.
\bibliography{anthology,custom}
\bibliographystyle{acl_natbib}

\appendix

\section{Hyperparameter Tuning}
\label{sec:hyper}
All hyperparameters were originally tuned with grid search on the development set. In this section, we show performance on the test set for better analysis.

{\noindent \textbf{Keyword Selection rate $p$}}: The keywords in our template provide the model with contextually relevant additional knowledge about the NEs during training and generation. However, we are faced with the question: \emph{How much context is good context?}. Too less context, like our ACLM \emph{only entity} baseline with only linearized NEs in the template, might make it difficult for the model to know the appropriate context of the syntactically ambiguous complex NE and thus might lead to sentences generated with a context-entity mismatch (for e.g. \emph{sam is reading on the Beach} where \emph{on the beach} might be a name of a movie). On the contrary, retaining too many words from the original sentence in our template might lead to a drop in the diversity of generated sentences as the model needs to \emph{infill} only a small portion of the words. To determine the optimal value of $p$ we experiment on 2 low-resource settings on the English sub-set of MultiCoNER and report the micro F1 results on the test-set for $p$ $\in$ \{0, 0.1, 0.2, 0.3. 0.4, 0.5, 0.6, 0.7\}. All other hyper-parameters are kept constant. As shown in Table \ref{tab:performance_keyword}, $p$ = 0.3 gives us the best test-set performance, and the performance decreases after 0.4.

\begin{table}[h]
    \centering
    \resizebox{0.9\columnwidth}{!}{
    \begin{tabular}{ccccccccc}
    \hline\hline
        \textbf{\#Gold} & \textbf{0} & \textbf{0.1} & \textbf{0.2} & \textbf{0.3} & \textbf{0.4} & \textbf{0.5} & \textbf{0.6} & \textbf{0.7} \\
        \hline
        200  & 50.06 & 51.82 & 53.99 & \cellcolor{magenta!20}\textbf{54.99} & 51.05 & 54.28 & 52.16 & 54.34  \\
        500 & 57.55 & 56.12 & 57.93 & \cellcolor{magenta!20}\textbf{58.31} & 57.55 & 56.88 & 56.60 & 58.10 \\
         \hline
    \end{tabular}
    }
    \caption{Test set F1 for various Keyword Selection rates.}
    \label{tab:performance_keyword}
\end{table}

{\noindent \textbf{Augmentation rounds $R$:}} Augmenting the training dataset with several augmentation rounds $R$ proves effective until a saturation point is reached. Continuing to add more augmented data to the gold dataset starts introducing noise to the combined data. Additionally, with an increase in $R$, the chances of auto-regressive generation with \emph{top-k} sampling generating similar sentences increase. To determine the optimal value of $R$, we experiment on 2 low-resource settings on the English sub-set of MultCoNER and report the micro F1 results on the test-set for $R$ ranging from 1 to 7. All other hyperparameters are kept constant. As shown in Table 5, $R$ = 5 gives us the best test-set performance, and the performance decreases after 5 rounds.

\begin{table}[h]
    \centering
    \resizebox{0.9\columnwidth}{!}{
    \begin{tabular}{cccccccc}
    \hline\hline
        \textbf{\#Gold} & \textbf{1} & \textbf{2} & \textbf{3} & \textbf{4} & \textbf{5} & \textbf{6} & \textbf{7}\\
        \hline
        200  & 52.37 & 53.96 & 52.40 & 50.05 & \cellcolor{magenta!20}\textbf{54.99} & 53.46 & 53.75 \\
        500 & 58.24 & 58.17 & 57.90 & 58.11 & \cellcolor{magenta!20}\textbf{58.31} & 57.20 & 57.40  \\
         \hline
    \end{tabular}
    }
    \caption{Test set F1 for the number of augmentation rounds.}
    \label{tab:performance}
\end{table}
{\noindent \textbf{Attention layers $a$ for Keyword Selection:}} Selecting the right keywords for creating a template is integral to the success of ACLM. A clear example of this can be seen in Table \ref{tab:performance_mono_cross}, where ACLM outperforms ACLM \emph{random} (which chooses random tokens as keywords for template creation) by a significant margin. Transformer encoders consist of multiple layers, and each layer consists of multiple attention heads. While all heads in the same layer tend to behave similarly, different layers generally encode different semantic and syntactic information \cite{clark2019does}. Thus we experiment with different values of $\alpha$, or different combinations of transformer encoder layers which are used for calculating the attention scores for keyword selection. As mentioned in Section \ref{sec:template_gen}, by default, we average attention scores across all tokens, all heads, and the last $\alpha$ layers. For all our low-resource experiments, we use attention maps from a 24-layer XLMRoBERTa-large fine-tuned on the low-resource gold dataset for that particular setting. Table \ref{tab:attention_selection} compares the performance of 3 settings of $\alpha$ on 2 low-resource settings on the English sub-set of MultCoNER: \textbf{1.} Only last layer \textbf{2.} Last 4 layers. \textbf{3.} All 24 layers. As clearly evident, though setting 2 achieves the best performance, the difference in performance among different values of $\alpha$ is not too high. As part of future work, we would like to explore better ways to search for the optimal $\alpha$.

\begin{table}[h]
    \centering
    \tiny
    \resizebox{0.6\columnwidth}{!}{
    \begin{tabular}{cccc}
    \hline\hline
        \textbf{\#Gold} & \textbf{1} & \textbf{2} & \textbf{3}  \\
        \hline
        200  & 52.43 & \cellcolor{magenta!20}\textbf{54.99} & 54.13   \\
        500 &  58.09 & \cellcolor{magenta!20}\textbf{58.31} & 58.15  \\
         \hline
    \end{tabular}
    }
    \caption{Test set F1 for various settings of $\alpha$}
    \label{tab:attention_selection}
\end{table}


\section{Additional Results}
\label{subsec:train_all}
{\noindent \textbf{Current state of state-of-the-art:}} Most current state-of-the-art systems are built and evaluated on common NER benchmarks like CoNLL 2003 and OntoNotes v5.0. As discussed in Section \ref{sec:back_rel}, these benchmarks do not represent contemporary challenges in NER and contain sentences with easy entities and rich context. Table \ref{tab:performance_simple_ner} compares the performance of a simple XLM-R \cite{conneau2019unsupervised}, and Co-regularized LUKE \cite{zhou2021learning} (SOTA NER system) on 2 common NER and 1 complex NER benchmarks in both low- and high-resource settings. As we can clearly see, both systems achieve remarkable performance on both CoNLL 2003 and OntoNotes v5.0 but struggle on MultiCoNER. Additionally, the gap widens in low-resource settings. 
\begin{table}[h]
    \small
    \centering
    \resizebox{0.99\columnwidth}{!}{
    \begin{tabular}{clcc}
    \hline\hline
        \textbf{\#Gold} & \textbf{Method} & \textbf{XLM-R} & \textbf{Co-regularized LUKE} \\
        \hline
         & CoNLL 2003 & 84.82 & 86.92 \\
        500 & OntoNotes & 65.48 & 64.92 \\
        & MultiCoNER & 55.51 & 55.12  \\ 
         \hline
         & CoNLL 2003 & 92.21 & 92.56 \\
        All & OntoNotes & 85.07 & 87.57 \\
        & MultiCoNER & 70.31 & 69.58 \\
         \hline
    \end{tabular}
    }
    \caption{\small Performance comparison of XLM-RoBERTa \cite{conneau2019unsupervised} and Co-regularized LUKE \cite{zhou2021learning} on two common benchmark NER datasets and MultiCoNER \cite{malmasi2022multiconer} (complex NER benchmark) in both high- and low-resource settings. Co-regularized LUKE is the current SOTA NER system on both CoNLL 2003 and OntoNotes v5.0. \emph{Complex NER remains a difficult NLP task in both low- and high-resource labeled data settings.}}
    \label{tab:performance_simple_ner}
\end{table}

{\noindent \textbf{Training on the entire dataset:}} Beyond just evaluating ACLM performance on low-resource settings, we also compare ACLM with all our baselines on the entire MultiCoNER dataset (each language split contains $\approx$ 15300 sentences). Similar to low-resource settings, ACLM outperforms all our baselines across all languages and achieves an absolute average gain of 1.58\% over our best baseline.

\begin{table}[h]
\centering
\resizebox{\columnwidth}{!}{
\begin{tabular}{llllllllllll}
\hline\hline
\textbf{Method} & \textbf{En} & \textbf{Bn} & \textbf{Hi} & \textbf{De} & \textbf{Es} & \textbf{Ko }& \textbf{Nl} & \textbf{Ru} & \textbf{Tr} & \textbf{Zh}& \textbf{Avg}  \\ 
\hline
Gold-only & 71.25 & 59.10 & 61.59 & 75.33 & 67.71 & 65.29 & 71.55 & 68.76 & 62.44 & 60.56 & 66.36    \\
LwTR   & 71.22   &  58.86  &  60.72  &  75.50  & 70.06   &  65.80  & 72.94   &  68.26  & 62.70   &  58.74  & 66.48 \\
DAGA   &  64.30  & 47.93   &   53.03 &  67.70  & 62.07   &  59.84  & 65.37  & 60.72  &  52.45 & 55.32 & 58.87    \\
 MELM                &66.27    & 56.27   &61.04    &71.25    & 65.56   & 63.71   &70.43    & 66.28 & 60.74   & 57.72   &  63.93 \\
\textbf{ACLM \emph{(ours)}}  & \cellcolor{magenta!20}\textbf{72.69} & \cellcolor{magenta!20}\textbf{60.13} & \cellcolor{magenta!20}\textbf{62.58} & \cellcolor{magenta!20}\textbf{77.26} & \cellcolor{magenta!20}\textbf{70.89} & \cellcolor{magenta!20}\textbf{67.01} & \cellcolor{magenta!20}\textbf{73.28} &  \cellcolor{magenta!20}\textbf{69.90}  &  \cellcolor{magenta!20}\textbf{65.24}  & \cellcolor{magenta!20}\textbf{61.63} & \cellcolor{magenta!20}\textbf{68.06} \\
\hline
\end{tabular}
}
\caption{\small Result comparison Complex NER. \textbf{Avg} is the average result across all languages. ACLM outperforms all our baselines.}
\label{tab:multi}
\end{table}


{\noindent \textbf{Entity-wise Performance Analysis:}} Previous to MultiCoNER, common benchmark datasets like CoNLL 2003 had only ``easy entities'' like names of Persons, Locations, and Organizations. The MultiCoNER dataset has 3 additional types of NEs, namely Products (\textbf{PROD}), Groups (\textbf{GRP}), and Creative Work (\textbf{CW}). These entities are syntactically ambiguous, which makes it challenging to recognize them based on their context. The top system from WNUT 2017 achieved 8\% recall for creative work entities. Table \ref{tab:entity_wise} compares the entity-wise performance of ACLM with our various baselines on two low-resource settings on the MultiCoNER dataset. All results are averaged across all 10 languages. ACLM outperforms all our baselines on all individual entities, including PROD, GRP, and CW, which re-affirms ACLM's ability to generate effective augmentation for complex NER.

\begin{table}[h]
    \tiny
    \centering
    \resizebox{0.99\columnwidth}{!}{
    \begin{tabular}{clcccccc}
    \hline\hline
        \textbf{\#Gold} & \textbf{Method} & \textbf{PER} & \textbf{LOC} & \textbf{PROD} & \textbf{GRP} & \textbf{CORP} & \textbf{CW}\\
        \hline
         \multirow{5}{*}{200} & Gold-Only & 56.35 & 42.32 & 30.10 & 31.36 & 33.83 & 23.30\\
          & LwTR & 56.13 & 41.78 & 34.87 & 36.52 & 39.30 & 27.46 \\
         & DAGA & 45.19 & 35.40 & 19.96 & 21.92 & 19.60 & 14.33 \\
          & MELM & 52.16 & 41.16 & 30.24 & 28.61 & 34.13 & 22.77\\
         & \textbf{ACLM \emph{(ours)}} & \cellcolor{magenta!20}\textbf{64.42} & \cellcolor{magenta!20}\textbf{48.92}  & \cellcolor{magenta!20}\textbf{41.76} & \cellcolor{magenta!20}\textbf{37.31} & \cellcolor{magenta!20}\textbf{44.08} & \cellcolor{magenta!20}\textbf{30.61} \\ 
         \hline
         \multirow{5}{*}{500}& Gold-Only & 63.05 & 48.48 & 42.75 & 37.55 & 45.10 & 31.34\\
         & LwTR & 64.80 & \cellcolor{magenta!20}\textbf{54.17} & 45.70 & \cellcolor{magenta!20}\textbf{44.06} & 50.80 & 35.10\\
         & DAGA & 51.82 & 41.11 & 28.58 & 30.50 & 34.10 & 21.61\\
         & MELM & 58.41 & 45.64 & 37.04 & 34.11 & 40.42 & 28.33\\
        & \textbf{ACLM \emph{(ours)}} & \cellcolor{magenta!20}\textbf{66.49} & 51.24  & \cellcolor{magenta!20}\textbf{48.87} & 42.00 & \cellcolor{magenta!20}\textbf{51.55} & \cellcolor{magenta!20}\textbf{35.18}\\
         \hline
    \end{tabular}
    }
    \caption{\small Entity-wise performance comparison of different augmentation methods. Results are averaged across all languages.}
    \label{tab:entity_wise}
\end{table}

{\noindent \textbf{Length-wise Performance Analysis:}} As mentioned in Section \ref{sec:back_rel}, low-context is a major problem in complex NER, and an effective complex NER system should be able to detect NEs in sentences with both low and high context (by context we refer to the number of words around the NEs in the sentence). By the nature of its fine-tuning pipeline, ACLM is able to generate augmentations of variable length, and our dynamic masking step further boosts the length diversity of generated augmentations. Adding to this, we acknowledge that effective augmentations for syntactically complex entity types should enable a model to learn to detect these entities in even low-context. Table \ref{tab:length_wise} compares the entity-wise performance of ACLM with our various baselines on two low-resource settings on the MultiCoNER dataset. All results are averaged across all 10 languages. ACLM outperforms all our baselines across all length settings, which re-affirms ACLM's ability to generate effective augmentation for complex NER. To be specific, ACLM improves over our best baseline by 8.8\% and 7.4\% for 200 and 3.2\% and 6.7\% for 500 for low- and high-context sentences, respectively.

\begin{table}[h]
    \tiny
    \centering
    \resizebox{0.99\columnwidth}{!}{
    \begin{tabular}{clccc}
    \hline\hline      
    \textbf{\#Gold} & \textbf{Method} & \textbf{len $<$ 5} & \textbf{ 5 $\leq$ len $<$ 10} & \textbf{10 $\leq$ len} \\
        \hline
         \multirow{4}{*}{200} & LwTR & 26.35 & 34.38 & 43.56\\
         & DAGA & 18.20 & 29.53 & 39.49  \\
          & MELM & 23.27 & 38.81 & 50.29 \\
         & \textbf{ACLM \emph{(ours)}} & \cellcolor{magenta!20}\textbf{35.10} & \cellcolor{magenta!20}\textbf{47.25} & \cellcolor{magenta!20}\textbf{57.72} \\ 
         \hline
         \multirow{4}{*}{500} & LwTR & 34.04 & 42.74 & 56.47 \\
         & DAGA & 23.00 & 38.18 & 51.09 \\
         & MELM & 27.46 & 44.74 & 57.91 \\
        &\textbf{ACLM \emph{(ours)}} & \cellcolor{magenta!20}\textbf{37.42} & \cellcolor{magenta!20}\textbf{52.23} & \cellcolor{magenta!20}\textbf{63.13}\\
         \hline
    \end{tabular}
    }
    \caption{\small Length-wise performance comparison of different augmentation methods. Results are averaged across all languages. ACLM outperforms all our baselines across all settings.}
    \label{tab:length_wise}
\end{table}

\begin{figure*}
    \centering
    \includegraphics[width=0.9\textwidth]{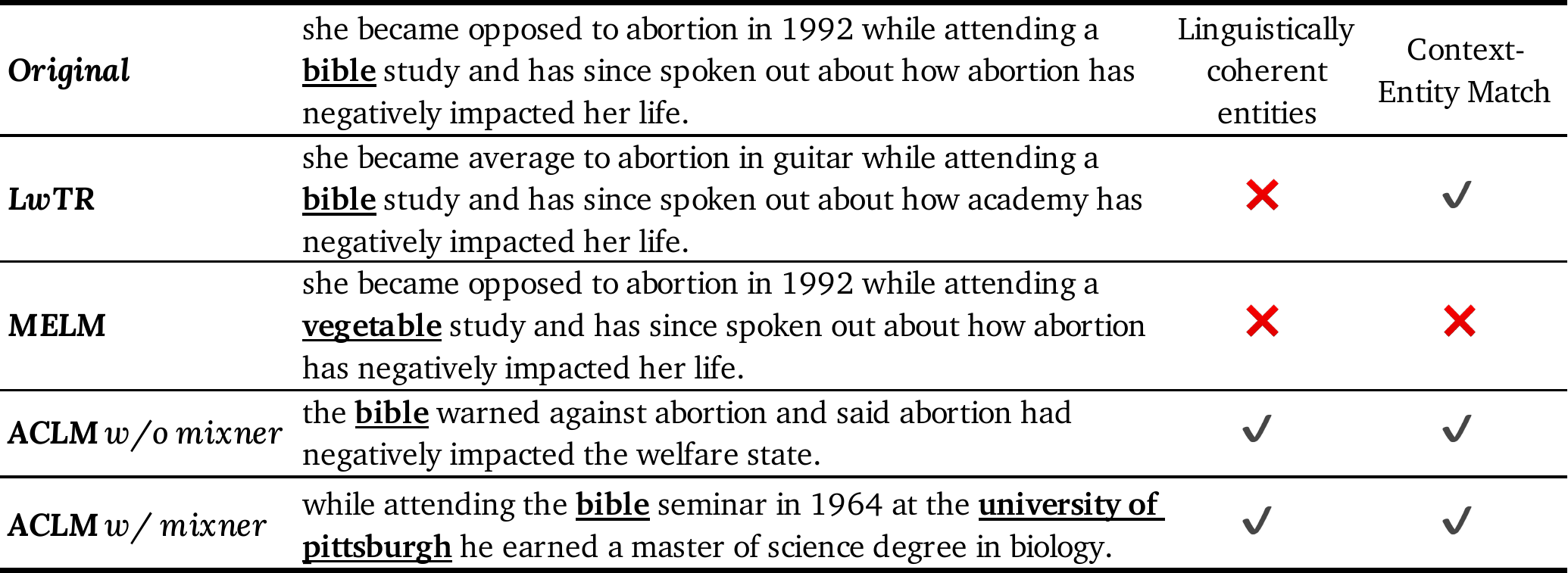}
    \caption{\small{Analysis and comparison of augmentations generated by our baselines with ACLM. Words \underline{\textbf{underlined}}} are the NEs. Context entity mismatch occurs when the generated NEs do not fit the surrounding context. Linguistic incoherence refers to cases where a generated NE does not follow the linguistic pattern for that particular type of NE or context.}
    \label{fig:linguistic}
\end{figure*}


\section{Templates and Attention Maps}
\label{sec:templates}
\begin{figure*}
    \includegraphics[width=1\textwidth]{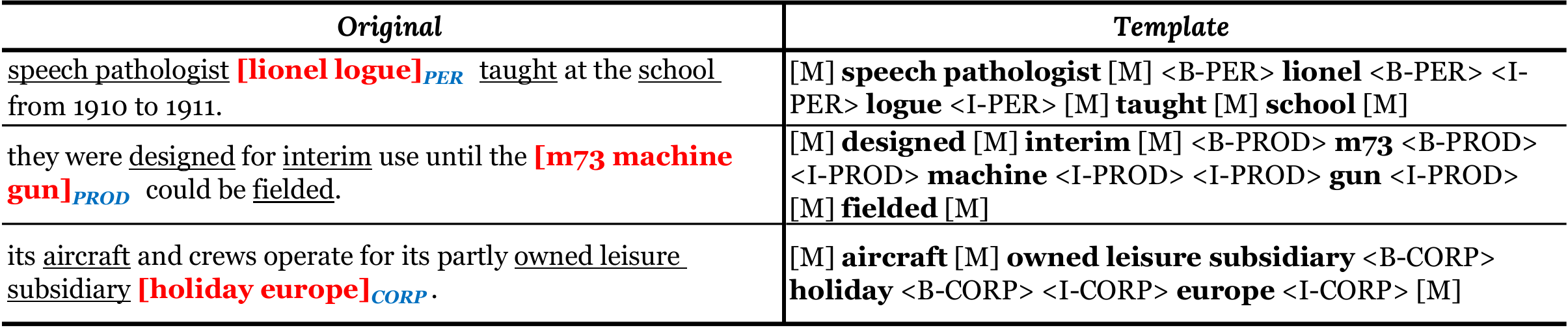}
    \caption{\small{Examples of templates created for sentences taken from the English subset of the MultiCoNER dataset. All templates shown are created in a low-resource setting (500 training examples). Words underlined are identified \emph{keywords}.}}
    \label{fig:eng2}
\end{figure*}

\begin{figure*}
    \includegraphics[width=1\textwidth]{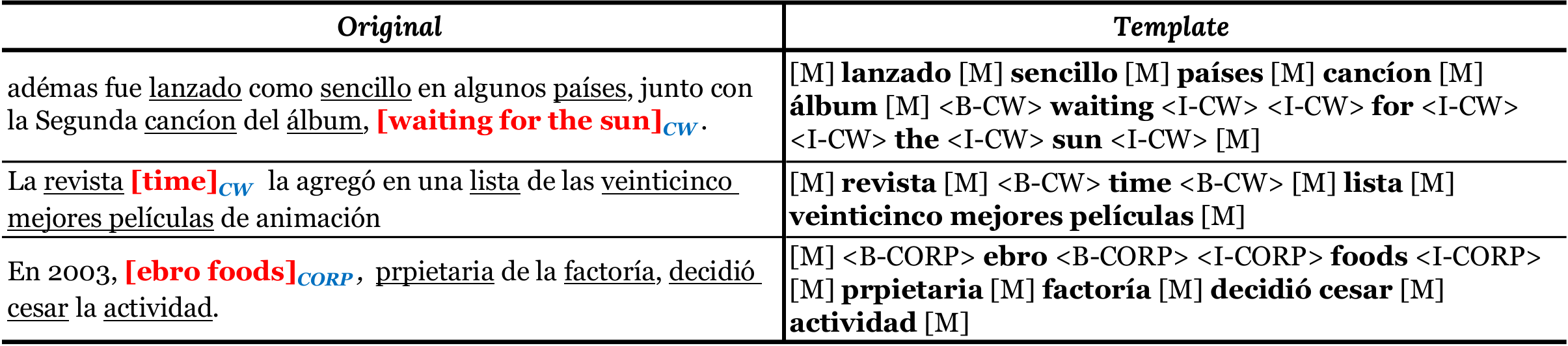}
    \caption{\small{Examples of templates created for sentences taken from the Spanish subset of the MultiCoNER dataset. All templates shown are created in a low-resource setting (500 training examples). Words underlined are identified \emph{keywords}.}}
    \label{fig:span2}
\end{figure*}

\begin{figure*}
    \includegraphics[width=1\textwidth]{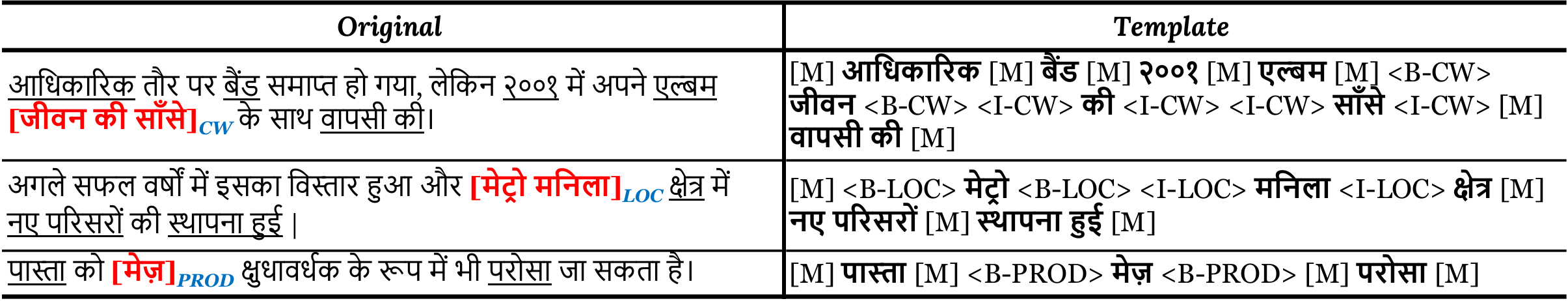}
    \caption{\small{Examples of templates created for sentences taken from the Hindi subset of the MultiCoNER dataset. All templates shown are created in a low-resource setting (500 training examples). Words underlined are identified \emph{keywords}.}}
    \label{fig:hindi2}
\end{figure*}

\begin{figure*}[]
\centering
\begin{subfigure}{0.4\textwidth}
\includegraphics[width=\textwidth]{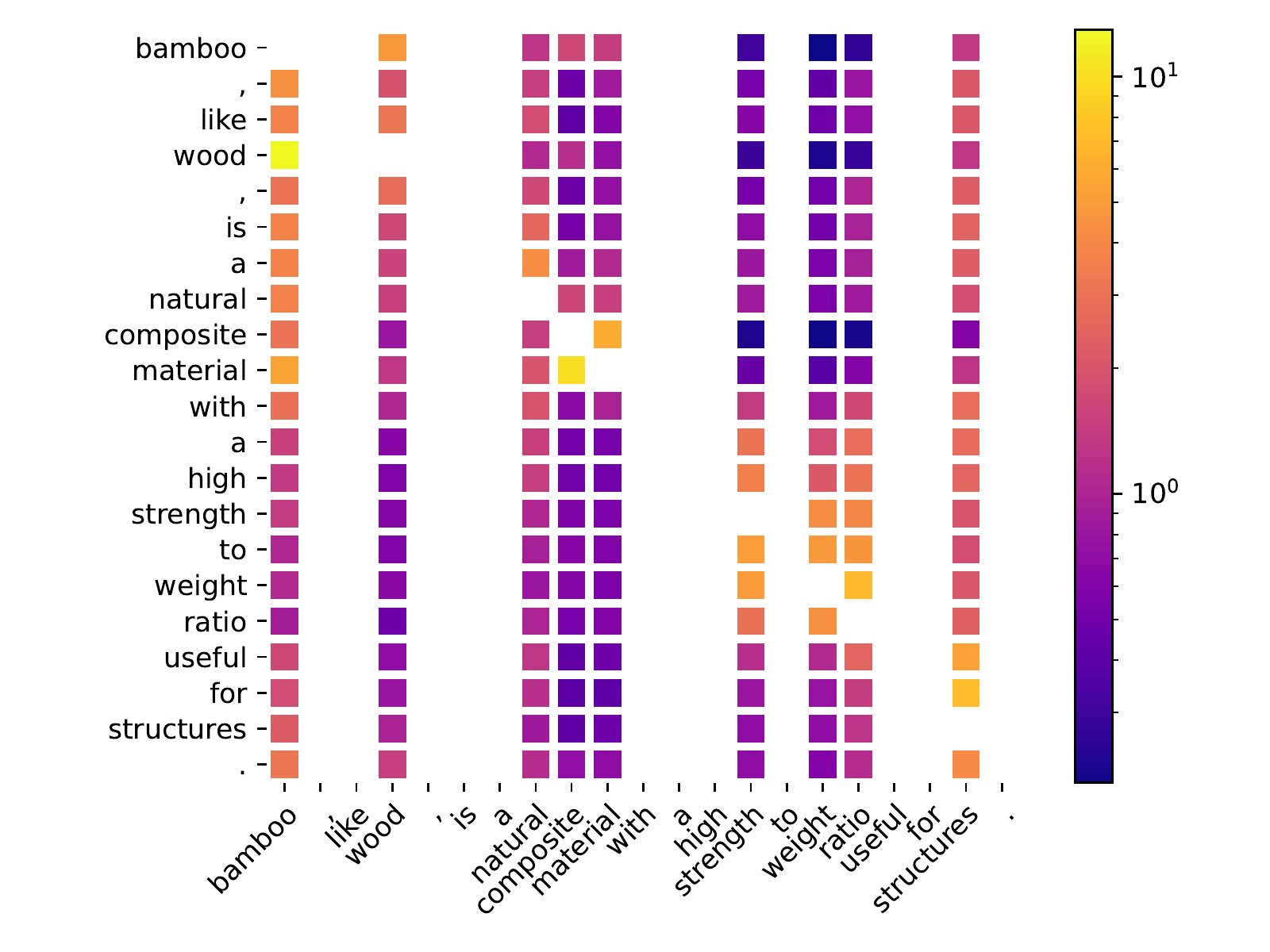}
\caption{Sentence: \underline{bamboo}, like \textcolor{red}{\textbf{[wood]}}\textcolor{blue}{\textsubscript{PROD}} is a \underline{natural} \textcolor{red}{\textbf{[composite material]}}\textcolor{blue}{\textsubscript{PROD}} with a high \underline{strength} to \underline{weight} \underline{ratio} useful for \underline{structures}.}
\label{fig:attention1}
\end{subfigure}
\hspace{30pt}
\begin{subfigure}{0.4\textwidth}
\includegraphics[width=\textwidth]{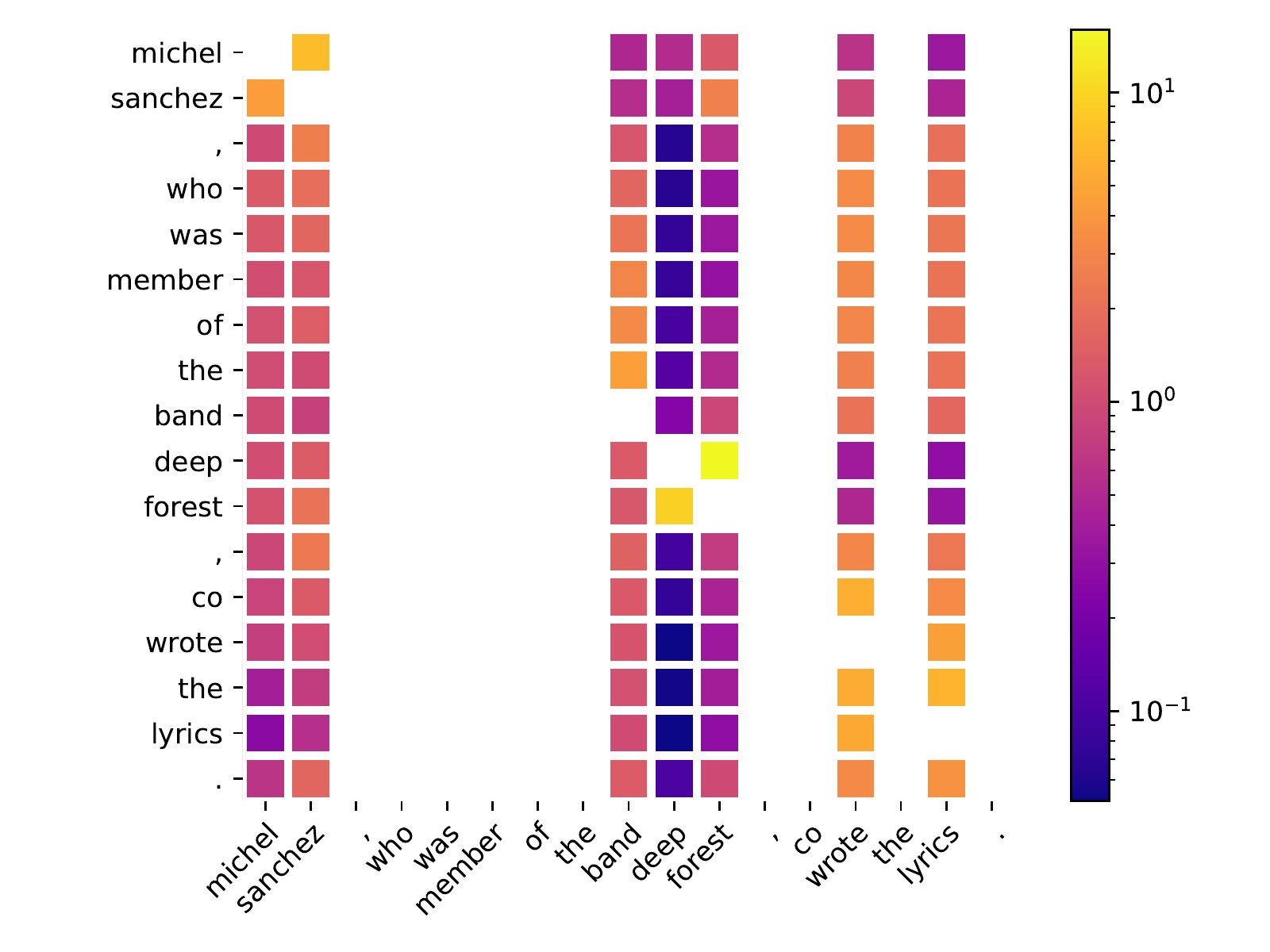}
\caption{Sentence: \textcolor{red}{\textbf{[michael sanchez]}}\textcolor{blue}{\textsubscript{PER}}, who was member of the \underline{band} \textcolor{red}{\textbf{[deep forest]}}\textcolor{blue}{\textsubscript{GRP}}, co \underline{wrote} the \underline{lyrics}.}
\label{fig:attention2}
\end{subfigure}
\vspace{10pt}
\begin{subfigure}{0.4\textwidth}
\includegraphics[width=\textwidth]{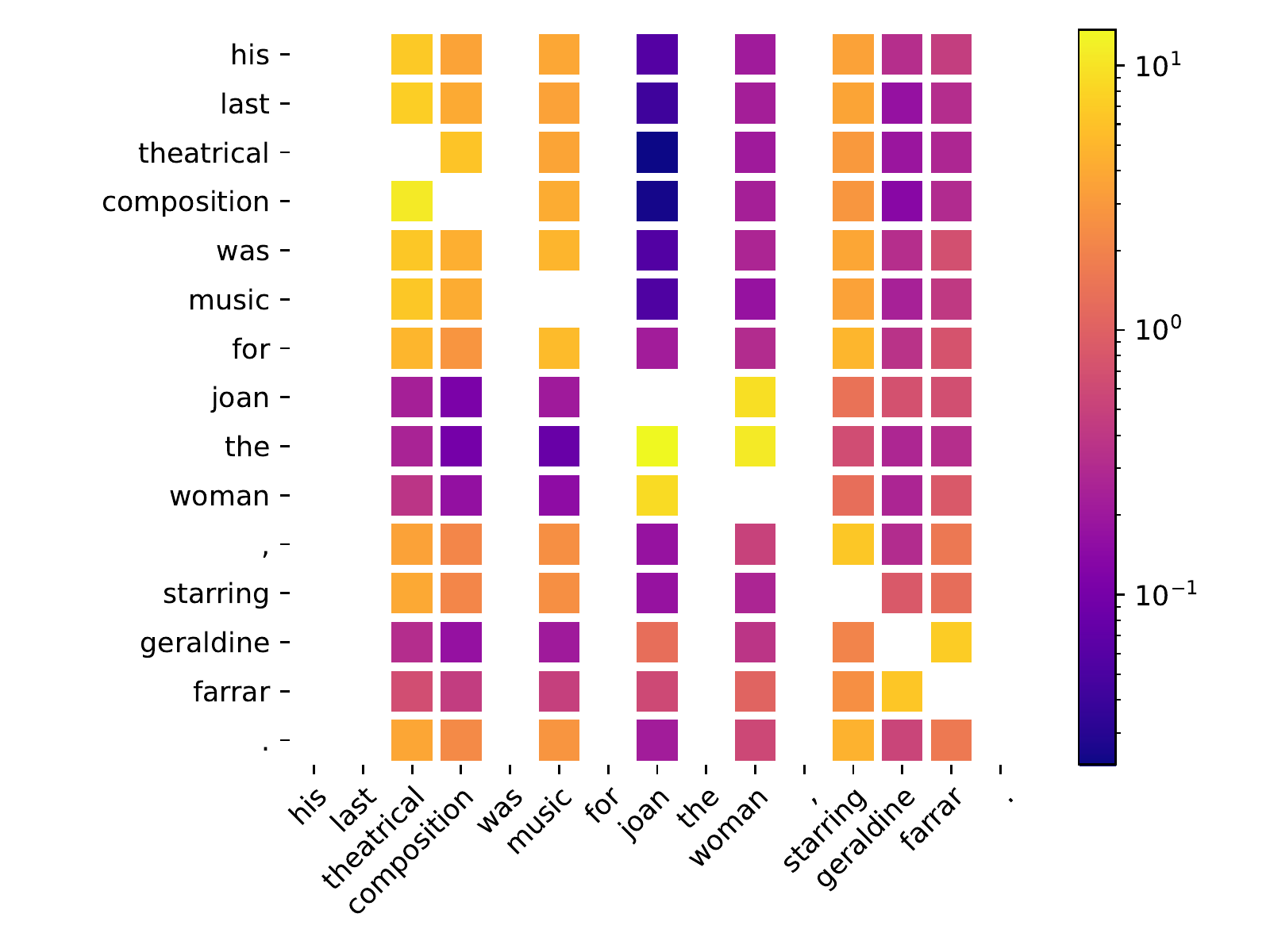}
\caption{Sentence: his last \underline{theatrical} composition was \underline{music} for \textcolor{red}{\textbf{[joan the woman]}}\textcolor{blue}{\textsubscript{CW}} starring \textcolor{red}{\textbf{[geraldine farrar]}}\textcolor{blue}{\textsubscript{PER}}.}
\label{fig:attention3}
\end{subfigure}
\hspace{30pt}
\begin{subfigure}{0.4\textwidth}
\includegraphics[width=\textwidth]{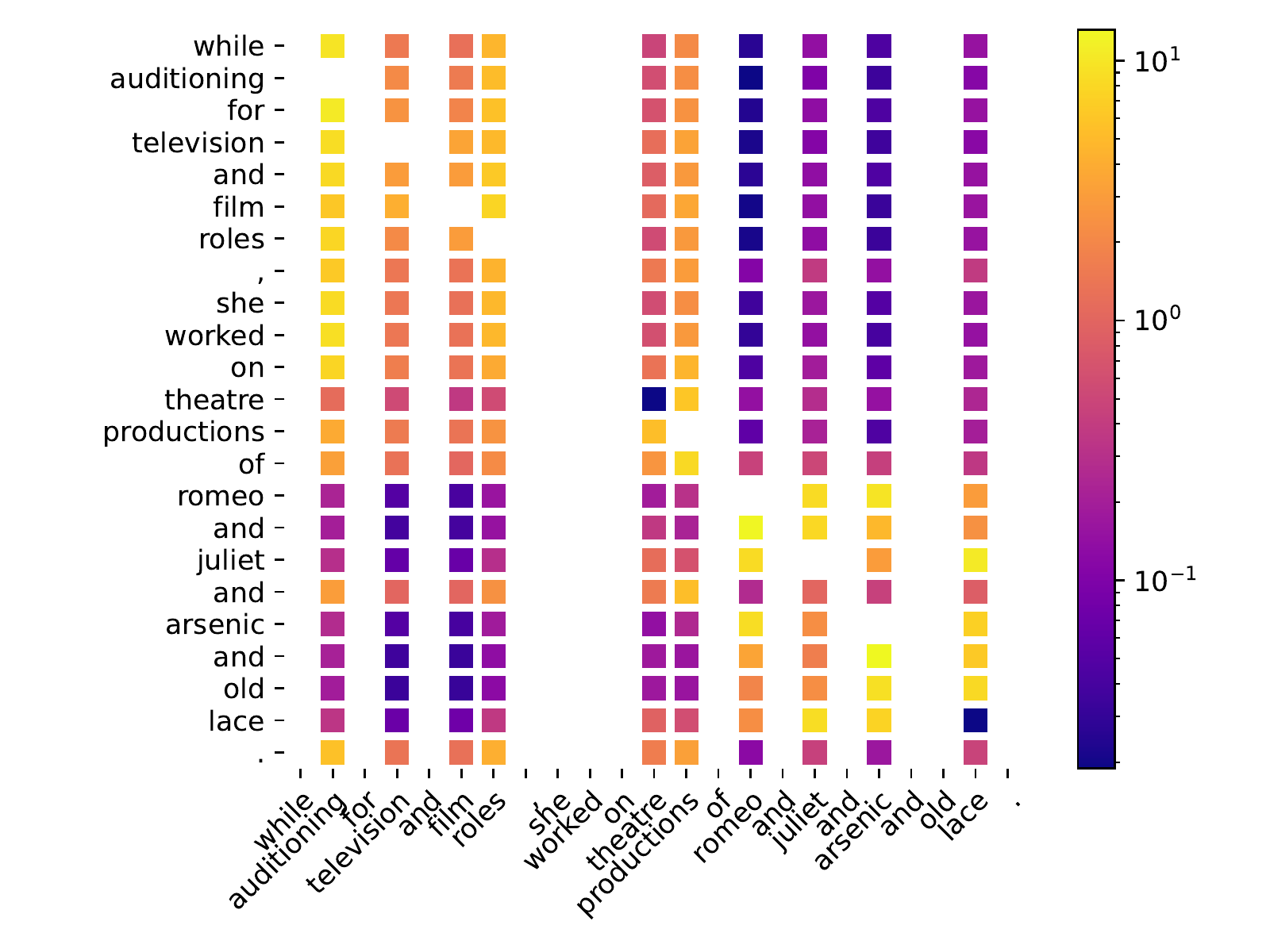}
\caption{Sentence: while \underline{auditioning} for television and film \underline{roles}, she worked on \textcolor{red}{\textbf{[theatre]}}\textcolor{blue}{\textsubscript{GRP}} \underline{productions} of \textcolor{red}{\textbf{[romeo and juliet]}}\textcolor{blue}{\textsubscript{CW}} and \textcolor{red}{\textbf{[arsenic and old lace]}}\textcolor{blue}{\textsubscript{CW}}.}
\label{fig:attention5}
\end{subfigure}
\vspace{10pt}
\begin{subfigure}{0.4\textwidth}
\includegraphics[width=\textwidth]{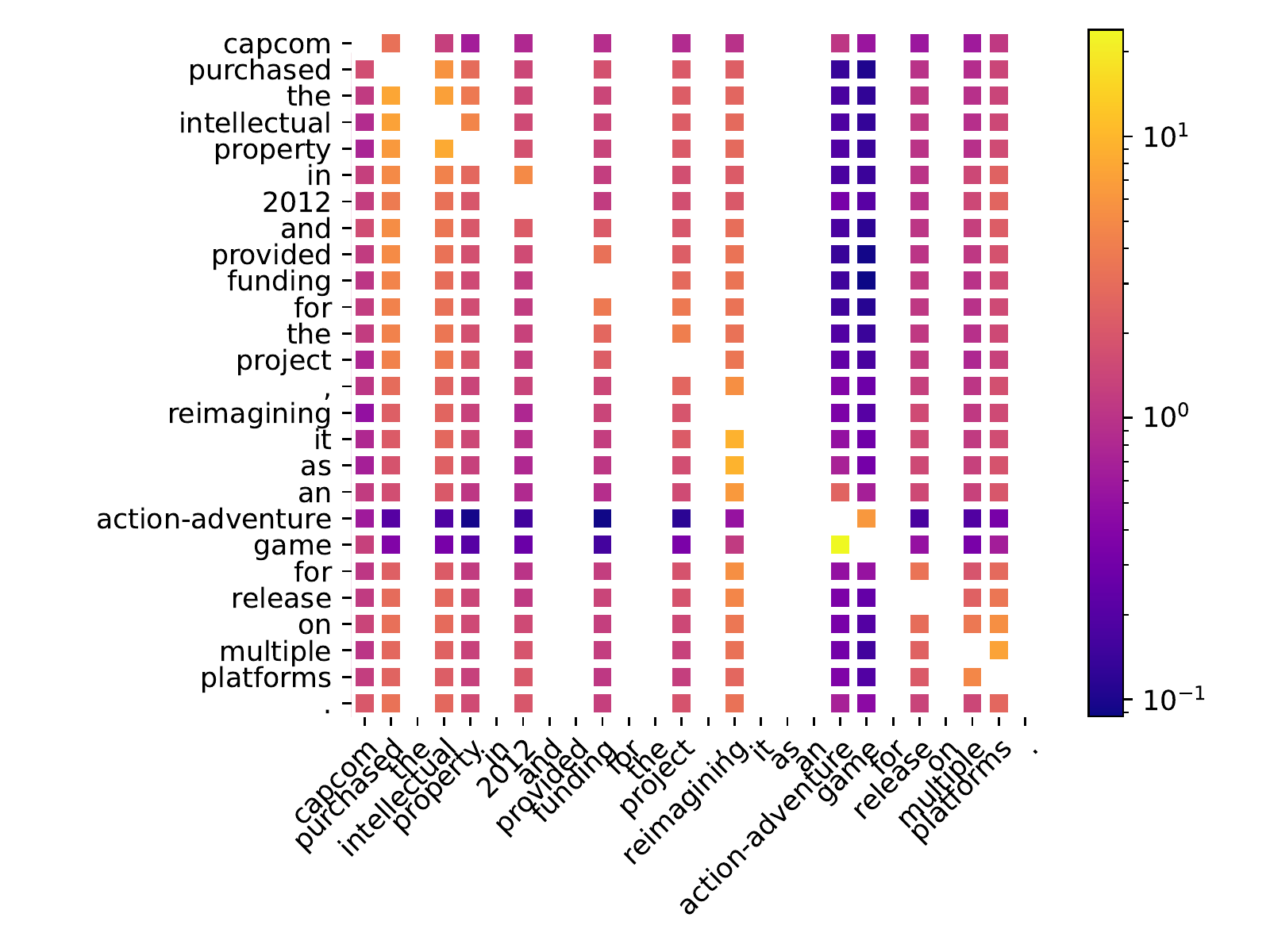}
\caption{Sentence: \textcolor{red}{\textbf{[capcom]}}\textcolor{blue}{\textsubscript{CORP}} \underline{purchased} the \underline{intellectual} \underline{property} in 2012 and provided \underline{funding} for the \underline{project} again,\underline{reimagining} it as an \textcolor{red}{\textbf{[action-adventure game]}}\textcolor{blue}{\textsubscript{CW}} for \underline{release} on multiple \underline{platforms}. }
\label{fig:attention4}
\end{subfigure}
\hspace{30pt}
\begin{subfigure}{0.4\textwidth}
\includegraphics[width=\textwidth]{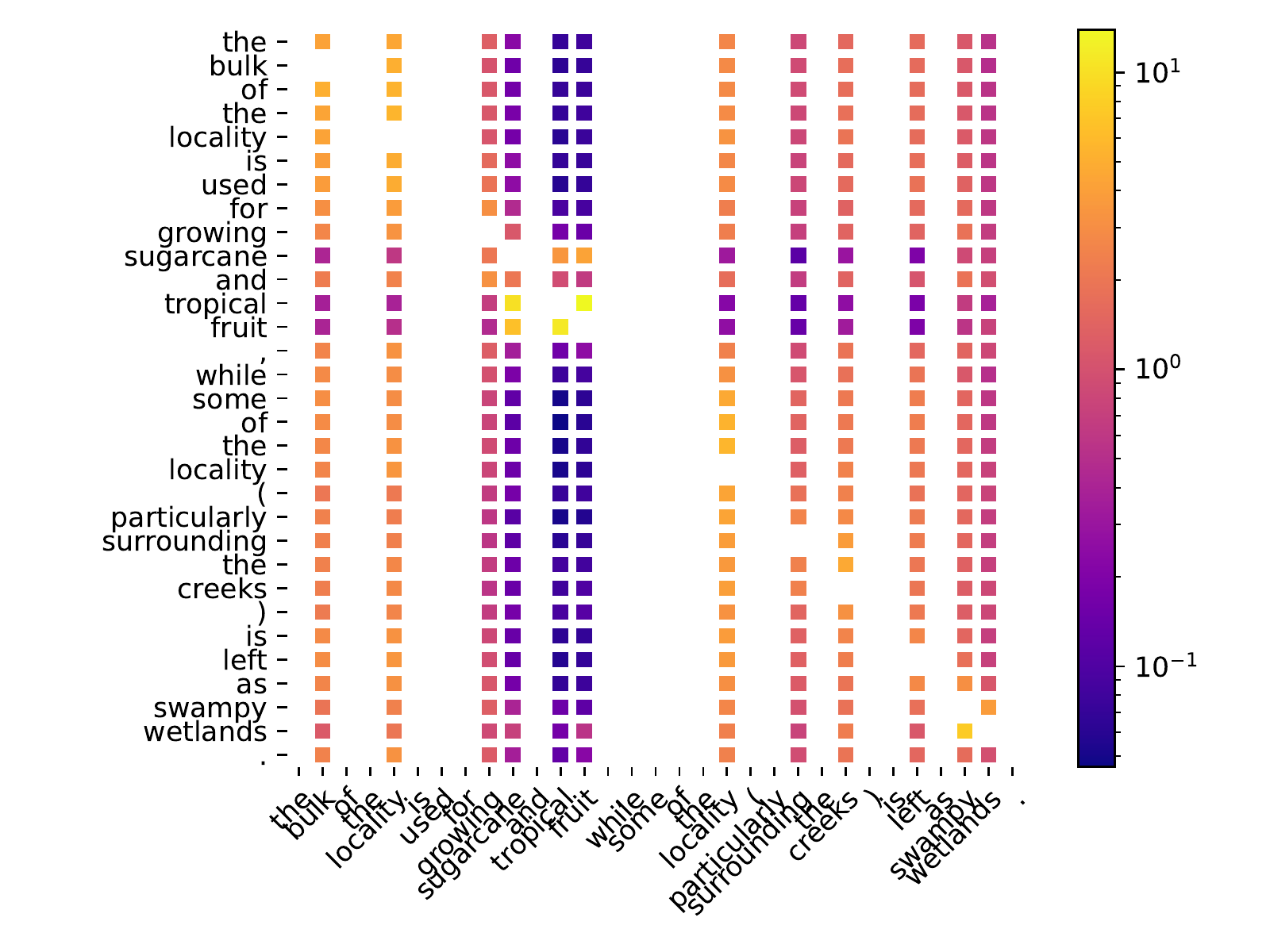}
\caption{Sentence: the \underline{bulk} of the \underline{locality} is used for \underline{growing} \textcolor{red}{\textbf{[sugarcane]}}\textcolor{blue}{\textsubscript{PROD}} and \underline{tropical} \underline{fruit}, while some of the \underline{locality}, \\ particularly surrounding  the \underline{creeks} is left as \underline{swampy} \underline{wetlands}. }
\label{fig:attention6}
\end{subfigure}
\caption{Attention maps for different sentences from the MULTICONER dataset. All the sentences are picked from a low-resource setting (1000 training examples).}
\end{figure*}

Creating templates with \emph{keywords} that effectively provides the PLM with additional knowledge about the NEs in the sentence is an integral part of ACLM. Fig. \ref{fig:eng2}, \ref{fig:span2}, \ref{fig:hindi2}, \ref{fig:ncb2}, \ref{fig:tdm2} shows examples of templates created for our sentences in MultiCoNER English subset, Spanish subset, Hindi subset NCBI Disease and TDMSci datasets, respectively. Additionally, we provide examples of attention maps used to create templates in Fig. \ref{fig:attention6}.

\section{Qualitative Analysis of Augmentations}
\label{sec:qual_analysis}

\subsection{Augmentation Examples}
\label{subsec:aug_examples}

\begin{figure*}
    \includegraphics[width=1\textwidth]{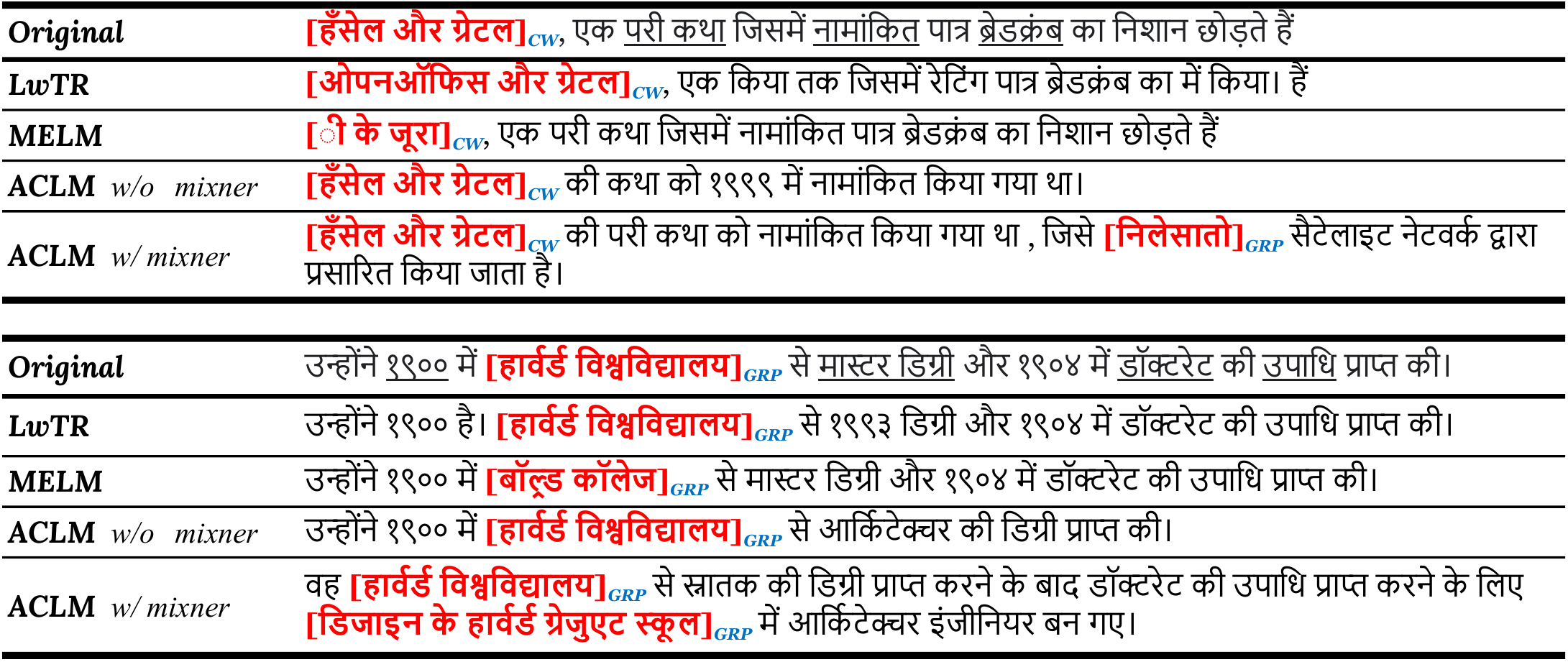}
    \caption{\small{Augmentation examples on the Hindi subset of the MultiCoNER dataset. All generations are produced in a low-resource setting (500 training examples).}}
    \label{fig:hi_1}
\end{figure*}

\begin{figure*}
    \includegraphics[width=1\textwidth]{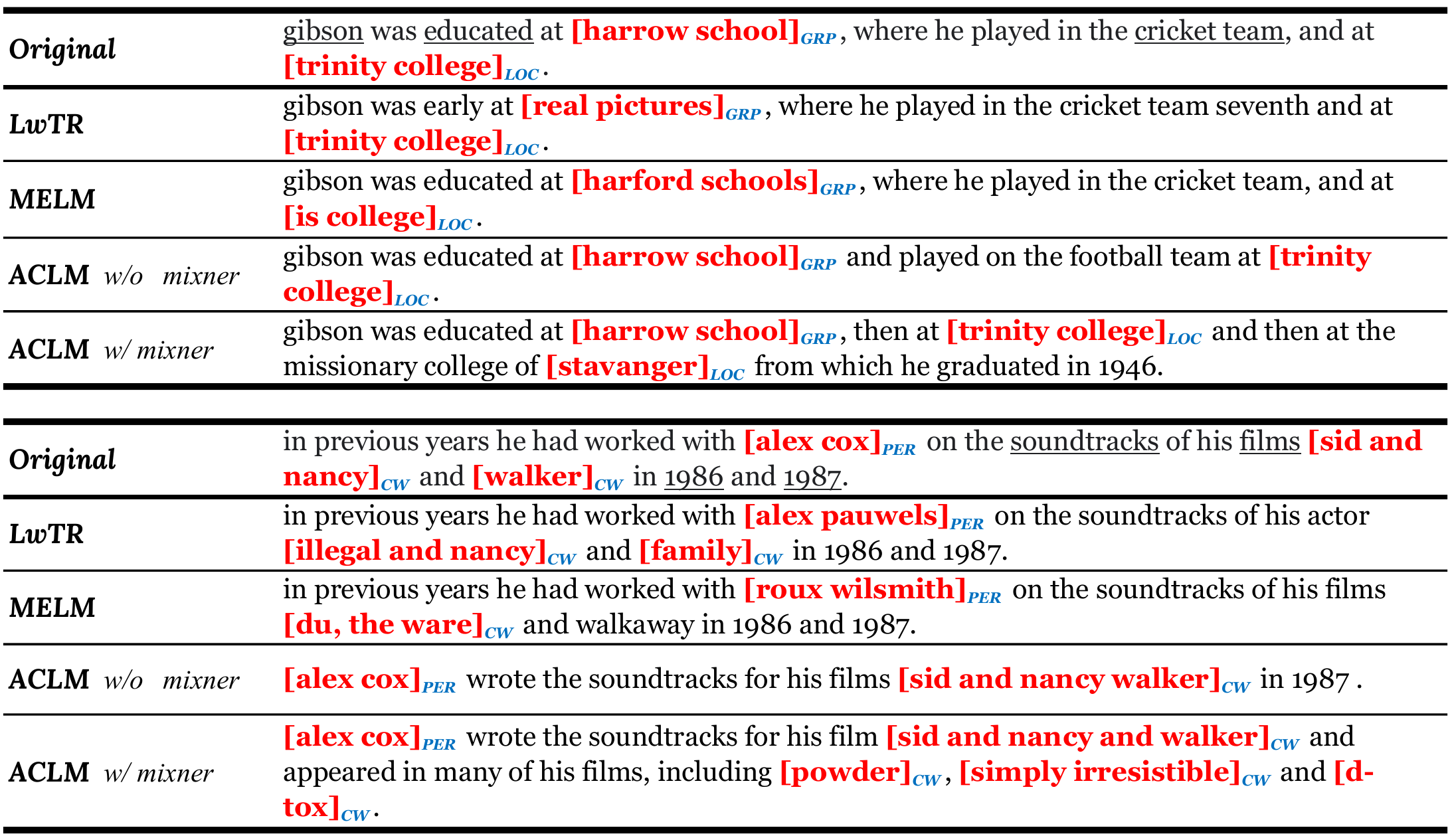}
    \caption{\small{Augmentation examples on the English subset of the MultiCoNER dataset. All generations are produced in a low-resource setting (500 training examples).}}
    \label{fig:eng_1}
\end{figure*}

{\noindent \textbf{MultiCoNER Dataset:}} We provide additional examples of augmentations generated by ACLM and all our baselines in Fig. \ref{fig:hi_1} and Fig. \ref{fig:eng_1} for Hindi and English subsets of MultiCoNER dataset respectively.

\begin{figure*}
    \includegraphics[width=1\textwidth]{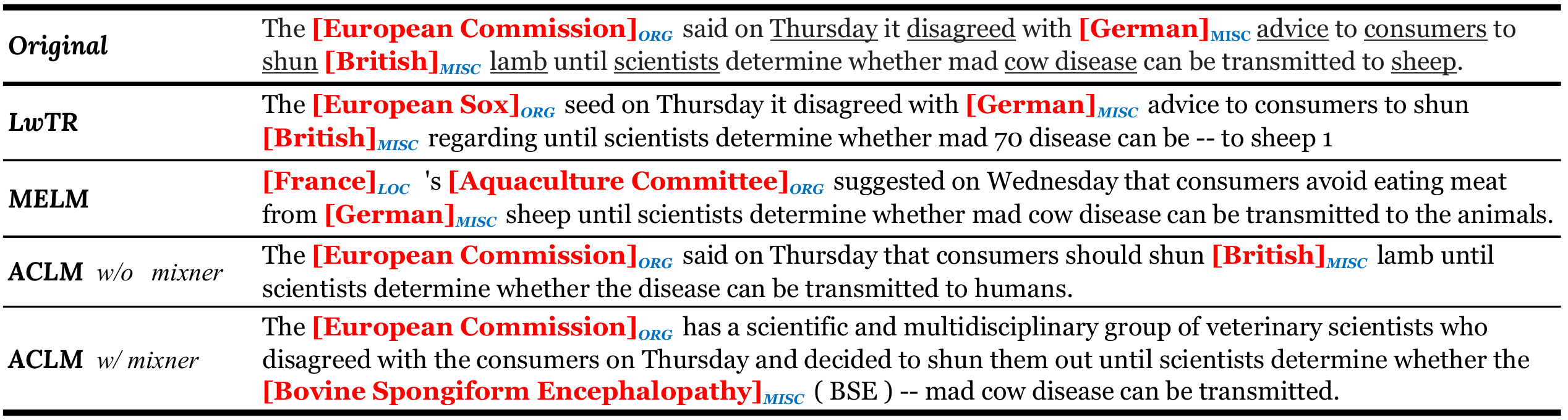}
    \caption{\small{Augmentation examples of the CoNLL 2003 dataset from the news domain. All generations are produced in a low-resource setting (500 training examples).}}
    \label{fig:conll}
\end{figure*}

\begin{figure*}
    \includegraphics[width=1\textwidth]{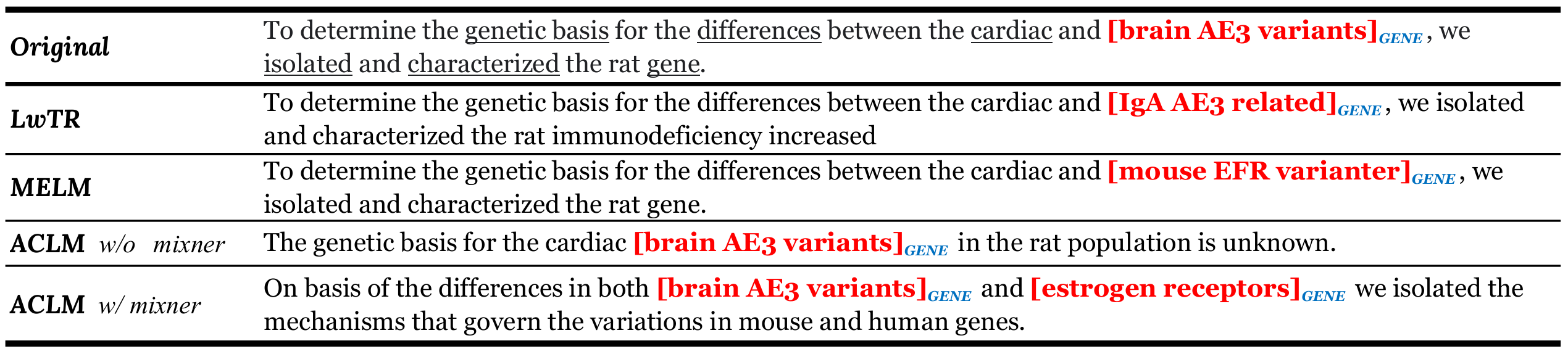}
    \caption{\small{Augmentation examples of BC2GM from the bio-medical domain. All generations are produced in a low-resource setting (500 training examples).}}
    \label{fig:bc2gm}
\end{figure*}

\begin{figure*}
    \includegraphics[width=1\textwidth]{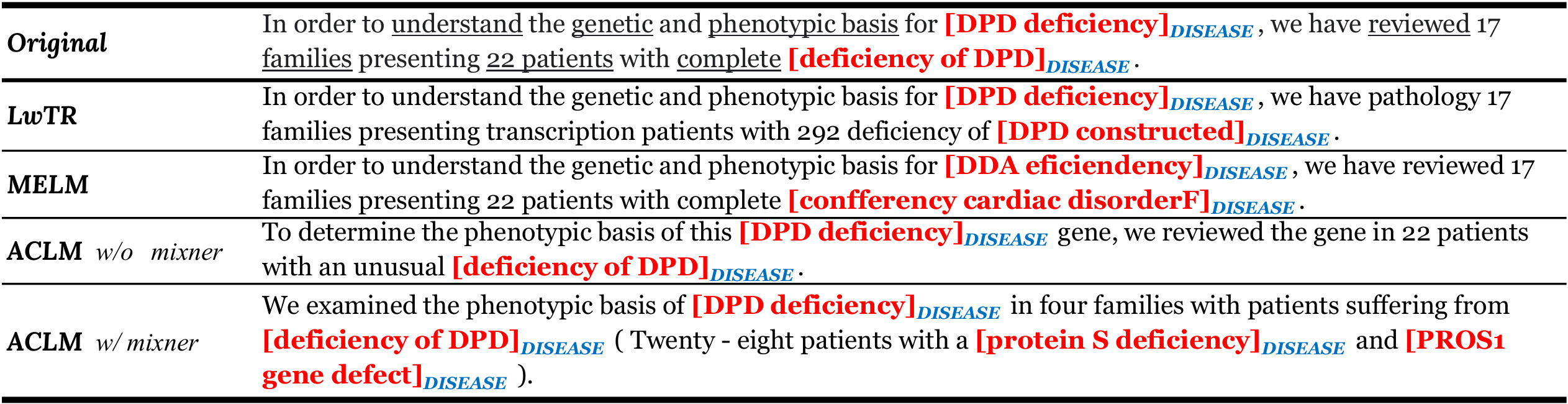}
    \caption{\small{Augmentation examples of NCBI dataset from the bio-medical domain. All generations are produced in a low-resource setting (500 training examples).}}
    \label{fig:ncbi}
\end{figure*}

\begin{figure*}
    \includegraphics[width=1\textwidth]{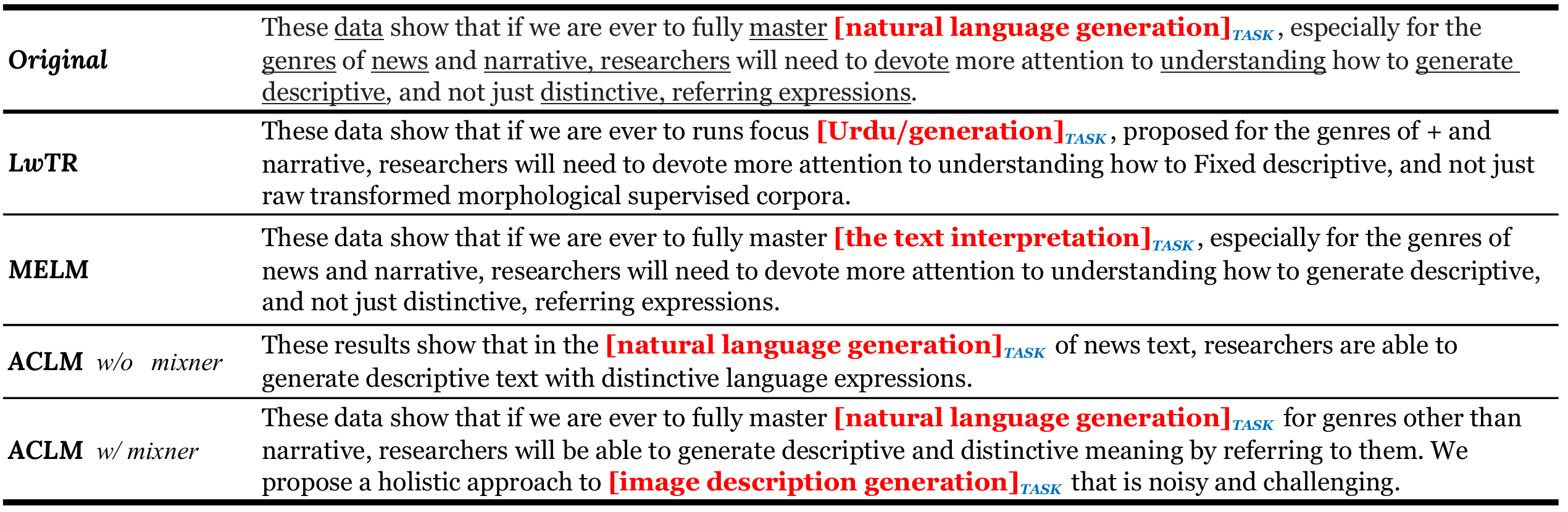}
    \caption{\small{Augmentation examples of TDMSci from the science domain. All generations are produced in a low-resource setting (500 training examples).}}
    \label{fig:tdmsci}
\end{figure*}

\begin{figure*}
    \includegraphics[width=1\textwidth]{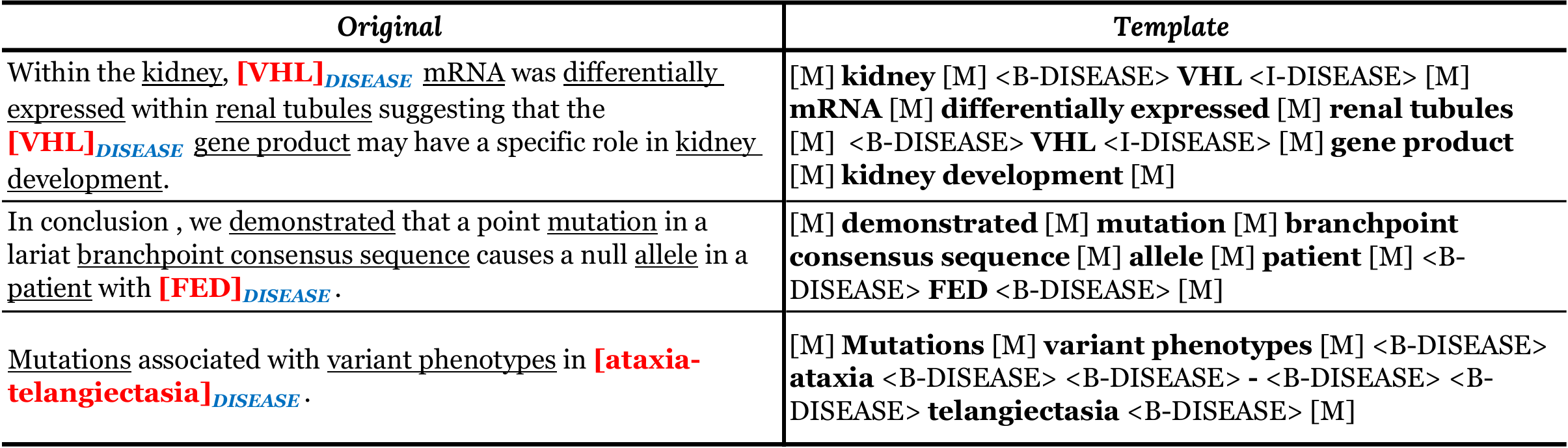}
    \caption{\small{Examples of templates created for sentences taken from the NCBI Disease dataset. All templates shown are created in a low-resource setting (500 training examples). Words underlined are identified \emph{keywords}.}}
    \label{fig:ncb2}
\end{figure*}

\begin{figure*}
    \includegraphics[width=1\textwidth]{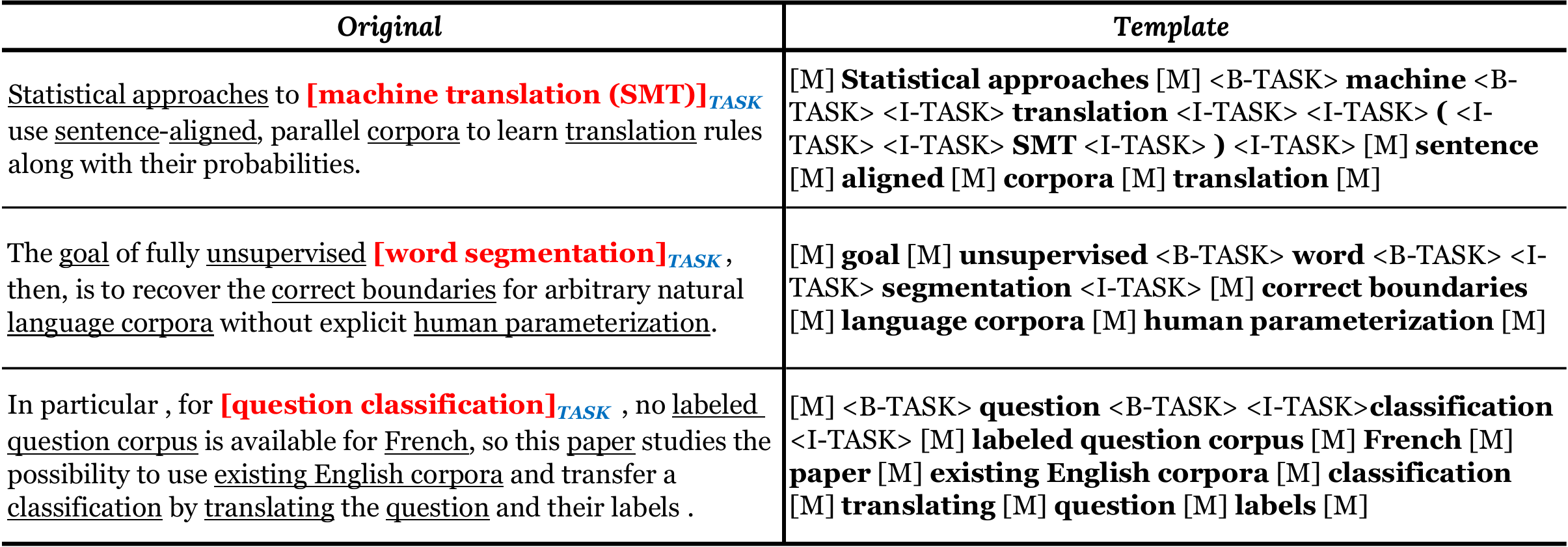}
    \caption{\small{Examples of templates created for sentences taken from the TDMSci dataset. All templates shown are created in a low-resource setting (500 training examples). Words underlined are identified \emph{keywords}.}}
    \label{fig:tdm2}
\end{figure*}

{\noindent \textbf{Extra Datasets:}} Fig \ref{fig:conll}, \ref{fig:bc2gm}, \ref{fig:ncbi} and \ref{fig:tdmsci} illustrate augmetation examples for CoNLL 2003 \cite{conll} (news), BC2GM \cite{smith2008overview} (bio-medical), NCBI Disease \cite{dougan2014ncbi} (bio-medical) and TDMSci \cite{houyufang2021eacl} (science) datasets respectively. Except for on CoNLL 2003 datasets, both our baselines, LwTR and MELM, generate incoherent and unreliable training samples for the other 2 datasets. We only compare ACLM with LwTR and MELM as these methods don't generate augmentations from scratch and modify existing sentences. We define unreliable sentences as sentences generated with an entity-context mismatch (eg. a NE describing a disease prone to cows is placed in the context of humans or vice-versa). Generating unreliable augmentations prove fatal in data-sensitive domains like bio-medical as it may make the model store wrongful knowledge. Our detailed analysis of generated augmentations shows that: \textbf{(1)} LwTR is prone to generating such incoherent sentences because it randomly samples entities from the corpus with the same tag for replacement. \textbf{(2)} MELM on the other hand, fine-tuned on a transformer-encoder-based PLM, gets to see the entire context of the sentence for generating a new NE. However, it does not learn to focus on particular keywords and tends to generate a new NE based on the broader context of the sentence (e.g., it does not learn to differentiate between human and cow diseases and generates a new NE based on the broader context of the sentence). \textbf{(3)} ACLM generates highly reliable samples by conditioning on templates with keywords related to the NE. We illustrate examples of such templates in Fig. \ref{fig:ncb2} and \ref{fig:tdm2}.

\begin{table*}[t!]
\centering
\tiny
\resizebox{0.9\textwidth}{!}{
    \begin{tabular}{l l r r r r r r r r r r r r r}
    \hline\hline
    class & split & \langid{EN} & \langid{DE} & \langid{ES} & \langid{RU} & \langid{NL} & \langid{KO} & \langid{FA} & \langid{ZH} & \langid{HI} & \langid{TR} & \langid{BN} & \langid{MULTI} & \langid{MIX}\\
\midrule
\multirow{3}{*}{\texttt{PER}} & train & 5,397 & 5,288 & 4,706 & 3,683 & 4,408 & 4,536 & 4,270 & 2,225 & 2,418 & 4,414 & 2,606 & 43,951 & 296 \\
 & dev & 290 & 296 & 247 & 192 & 212 & 267 & 201 & 129 & 133 & 231 & 144 & 2,342 & 96 \\
 & test & 55,682 & 55,757 & 51,497 & 44,687 & 49,042 & 39,237 & 35,140 & 26,382 & 25,351 & 26,876 & 24,601 & 111,346 & 19,313 \\[1ex]
\multirow{3}{*}{\texttt{LOC}} & train & 4,799 & 4,778 & 4,968 & 4,219 & 5,529 & 6,299 & 5,683 & 6,986 & 2,614 & 5,804 & 2,351 & 54,030 & 325 \\
 & dev  & 234 & 296 & 274 & 221 & 299 & 323 & 324 & 378 & 131 & 351 & 101 & 2,932 & 108 \\
 & test & 59,082 & 59,231 & 58,742 & 54,945 & 63,317 & 52,573 & 45,043 & 43,289 & 31,546 & 34,609 & 29,628 & 141,013 & 23,111 \\[1ex]
\multirow{3}{*}{\texttt{GRP}} & train & 3,571 & 3,509 & 3,226 & 2,976 & 3,306 & 3,530 & 3,199 & 713 & 2,843 & 3,568 & 2,405 & 32,846 & 248 \\
 & dev  & 190 & 160 & 168 & 151 & 163 & 183 & 164 & 26 & 148 & 167 & 118 & 1,638 & 75 \\
 & test & 41,156 & 40,689 & 38,395 & 37,621 & 39,255 & 31,423 & 27,487 & 18,983 & 22,136 & 21,951 & 19,177 & 77,328 & 16,357\\[1ex]
\multirow{3}{*}{\texttt{CORP}} & train & 3,111 & 3,083 & 2,898 & 2,817 & 2,813 & 3,313 & 2,991 & 3,805 & 2,700 & 2,761 & 2,598 & 32,890 & 294\\
 & dev  & 193 & 165 & 141 & 159 & 163 & 156 & 160 & 192 & 134 & 148 & 127 & 1,738 & 112\\
 & test & 37,435 & 37,686 & 36,769 & 35,725 & 35,998 & 30,417 & 27,091 & 25,758 & 21,713 & 21,137 & 20,066 & 75,764 & 18,478\\[1ex]
\multirow{3}{*}{\texttt{CW}} & train & 3,752 & 3,507 & 3,690 & 3,224 & 3,340 & 3,883 & 3,693 & 5,248 & 2,304 & 3,574 & 2,157 & 38,372 & 298\\
 & dev  & 176 & 189 & 192 & 168 & 182 & 196 & 207 & 282 & 113 & 190 & 120 & 2,015 & 102\\
 & test & 42,781 & 42,133 & 43,563 & 39,947 & 41,366 & 33,880 & 30,822 & 30,713 & 21,781 & 23,408 & 21,280 & 89,273 & 20,313\\[1ex]
\multirow{3}{*}{\texttt{PROD}} & train & 2,923 & 2,961 & 3,040 & 2,921 & 2,935 & 3,082 & 2,955 & 4,854 & 3,077 & 3,184 & 3,188 & 35,120 & 316\\
 & dev  & 147 & 133 & 154 & 151 & 138 & 177 & 157 & 274 & 169 & 158 & 190 & 1,848 & 117\\
 & test & 36,786 & 36,483 & 36,782 & 36,533 & 36,964 & 29,751 & 26,590 & 28,058 & 22,393 & 21,388 & 20,878 & 75,871 & 20,255\\
 \midrule
 \midrule
 \multirow{3}{*}{\#instances} & train & 15,300 & 15,300 & 15,300 & 15,300 & 15,300 & 15,300 & 15,300 & 15,300 & 15,300 & 15,300 & 15,300 &  168,300 & 1,500\\
 & dev & 800 & 800 & 800 & 800 & 800 & 800 & 800 & 800 & 800 & 800 & 800 & 8,800 & 500\\
 & test & 217,818 & 217,824 & 217,887 & 217,501 & 217,337 & 178,249 & 165,702 & 151,661 & 141,565 & 136,935 & 133,119 &  471,911 & 100,000 \\
 \bottomrule
    \end{tabular}}
    \caption{\small{MultiCoNER dataset statistics for the different languages for the train/dev/test splits. The bottom three rows show the total number of sentences for each language.}}
    \label{tab:data_stats_multiconner}
\end{table*}



\section{Additional Details}
\label{sec:additional}

{\noindent \textbf{Model Parameters:}} XLM-RoBERTa-large has $\approx$ 355M parameters with 24-layers of encoder, 1027-hidden-state, 4096 feed-forward hidden-state and 16-heads. mBART-50-large $\approx$ has 680M parameters with 12 layers of encoder, 12 layers of decoder, 1024-hidden-state, and 16-heads.
\vspace{1mm}

{\noindent \textbf{Compute Infrastructure:}} All our experiments are conducted on a single NVIDIA A100 GPU. An entire ACLM training pipeline takes $\approx$ 40 minutes.
\vspace{1mm}

{\noindent \textbf{Dataset Details:}} We use 5 datasets in total for our experiments: MultiCoNER \footnote{https://registry.opendata.aws/multiconer/} \cite{malmasi2022multiconer} (CC BY 4.0 licensed), CoNLL 2003 \footnote{https://huggingface.co/datasets/conll2003} \cite{conll} (Apache License 2.0), BC2GM \footnote{https://github.com/spyysalo/bc2gm-corpus} \cite{smith2008overview} (MIT License), NCBI Disease \footnote{https://huggingface.co/datasets/ncbidisease} \cite{dougan2014ncbi} (Apache License 2.0) and TDMSci \footnote{https://github.com/IBM/science-result-extractor} \cite{houyufang2021eacl} (Apache License 2.0). All the datasets are available to use for research purposes, and for our work, we use all these datasets intended for their original purpose, i.e., NER. MultiCoNER has data in 11 languages, including code-mixed and multi-lingual subsets. We experiment with 10 monolingual subsets discussed in Section \ref{sec:dataset} with appropriate reason for not experimenting on Farsi in our Limitations Section. According to the original papers of all 5 datasets used in the research, none of them contains any information that names or uniquely identifies individual people or offensive content.
\vspace{1mm}

{\noindent \textbf{Data statistics (train/test/dev splits):}} Detailed dataset statistics for MultiCoNER, CoNLL 2003, BC2GM, NCBI Disease and TDMSci can be found in Table \ref{tab:data_stats_multiconner} (language codes in Table \ref{tab:multi_conner_languages}), \ref{tab:conll_stats}, \ref{tab:bc2_stat}, \ref{tab:ncbi_dataset} and \ref{tab:tdmsci_stats} respectively.
\vspace{1mm}

{\noindent \textbf{Implementation Software and Packages:}} We implement all our models in PyTorch \footnote{https://pytorch.org/} and use the HuggingFace \footnote{https://huggingface.co/} implementations of mBART50 and XLM-RoBERTA (base and large). We use the FLAIR toolkit \cite{akbik2019flair} to fine-tune all our NER models. 
\vspace{1mm}

{\noindent \textbf{Potential Risks:}} Conditional Language Models used for Natural Language Generation often tend to \emph{hallucinate} \cite{ji2022survey} and potentially generate nonsensical, unfaithful or harmful sentences to the provided source input that it is conditioned on.
\vspace{1mm}

\begin{table}[ht!]
\centering
\resizebox{.6\columnwidth}{!}{
\begin{tabular}{ l l | l l | l l }
\hline\hline
 Bangla &  (\langid{BN}) &  Hindi  & (\langid{HI})  & German  &  (\langid{DE}) \\ 
 Chinese &  (\langid{ZH}) & Korean  & (\langid{KO}) & Turkish & (\langid{TR}) \\  
 Dutch   &  (\langid{NL}) & Russian & (\langid{RU}) & Farsi   &  (\langid{FA})   \\
 English &  (\langid{EN}) & Spanish & (\langid{ES}) \\
\bottomrule
\end{tabular}}
\caption{\small The languages included in \mconer, along with their 2-letter codes.}
\label{tab:multi_conner_languages}
\end{table}

\begin{table}[ht!]
\centering
\resizebox{0.6\columnwidth}{!}{
\begin{tabular}{lccc}
\hline\hline
\multicolumn{1}{l}{ English data } & Articles & Sentences & Tokens \\
\hline Training set & 946 & 14,987 & 203,621 \\
Development set & 216 & 3,466 & 51,362 \\
Test set & 231 & 3,684 & 46,435 \\
\hline
\end{tabular}
}
    \caption{\small CoNLL Dataset Stats}
    \label{tab:conll_stats}
\end{table}

\begin{table}[h]
    \centering
\resizebox{0.35\columnwidth}{!}{
\begin{tabular}{lcc} 
\hline
\hline
& Train & Test \\
\hline
\# Sentences & 1500 & 500 \\
\# Task & 1219 & 396 \\
\# Dataset & 420 & 192 \\
\# Metric & 536 & 174 \\
\hline
\end{tabular}
}
    \caption{\small TDMSci dataset statistics for the train/test splits.}
    \label{tab:tdmsci_stats}
\end{table}

\begin{table}[ht!]
    \centering
    \resizebox{0.7\columnwidth}{!}{
    \begin{tabular}{c c c c}
    \hline\hline
        & \textbf{Train} & \textbf{Dev} & \textbf{Test} \\
    \hline
     \# Sentences & 15197 & 3061 & 6325 \\ 
     \hline
    \end{tabular}
    }
    \caption{BC2GM Dataset Train/Dev/Test Split}
    \label{tab:bc2_stat}
\end{table}

\begin{table}[ht!]
    \centering
\resizebox{\columnwidth}{!}{
\begin{tabular}{lllll}
\hline\hline 
Corpus characteristics & Training set & Development set & Test set & Whole corpus \\
\hline PubMed citations & 593 & 100 & 100 & 793 \\
Total disease mentions & 5145 & 787 & 960 & 6892 \\
Unique disease mentions & 1710 & 368 & 427 & 2136 \\
Unique concept ID & 670 & 176 & 203 & 790 \\
\hline
\end{tabular}
}
    \caption{NCBI disease dataset statistics for the train/dev/test splits.}
    \label{tab:ncbi_dataset}
\end{table}


\end{document}